\documentclass[10pt,twocolumn,letterpaper]{article}

\PassOptionsToPackage{table}{xcolor}
\usepackage[pagenumbers]{cvpr} 

\usepackage{amsmath,amssymb}

\usepackage{graphicx}

\usepackage{multirow}
\usepackage{bm}
\usepackage{booktabs}
\usepackage{colortbl}
\usepackage{microtype}
\usepackage{makecell}
\usepackage{pifont}

\newcommand{\cmark}{\ding{51}} 
\newcommand{\xmark}{\ding{55}} 

\usepackage{tabularx}
\makeatletter
\@ifundefined{NC@find@C}{\newcolumntype{C}{>{\centering\arraybackslash}X}}{}
\makeatother

\definecolor{cvprblue}{rgb}{0.21,0.49,0.74}
\usepackage[pagebackref,breaklinks,colorlinks,allcolors=cvprblue]{hyperref}

\def\confName{CVPR}
\def\confYear{2027}

\title{TIDE-Flow: Fine-Grained Temporal Evidence and Structured Displacement Matching for Spike Optical Flow}

\author{Hanwen~Liang$^{1}$ \quad Wenxuan~Liu$^{2}$ \quad Wenxin~Huang$^3$ \quad Jingling~Yuan$^1$ \quad Zhaofei~Yu$^2$\\ 
\quad Xian~Zhong$^{1}$  \quad Tiejun~Huang$^2$ \\
$^1$ Wuhan University of Technology \quad 
$^2$ Peking University \quad
$^3$ Hubei University \\
}

\begin{document}
\maketitle

\begin{abstract}
Spike cameras capture high-speed motion through an integrate-and-fire mechanism, providing fine-grained temporal observations without exposure blur. However, their binary streams encode accumulated illumination rather than instantaneous intensity, making reliable motion estimation challenging. Existing spike-based methods exploit temporal aggregation to construct motion representations, but this process may weaken fine-grained firing patterns. Meanwhile, correlation-based estimators often rely on learned matching features without explicitly characterizing the spatial distribution and ambiguity of candidate displacements. To address these challenges, we propose \textbf{TIDE-Flow}, a spike-based optical flow estimation framework that jointly exploits Temporal Integration Evidence (TIE) and Displacement Evidence Encoding (DEE). TIE combines raw spike patterns, multi-scale firing statistics, temporal moments, and boundary-aware interval cues, preserving complementary temporal information before spatial feature compression. DEE further integrates structured candidate-grid features with explicit displacement statistics to characterize matching geometry and competing correspondences. Both components are incorporated into a difference-correlation recurrent refinement framework, enabling effective temporal motion modeling and correspondence estimation. Extensive experiments demonstrate the effectiveness of TIDE-Flow, with controlled ablations validating the contributions of temporal and displacement evidence modeling.
\end{abstract}

\section{Introduction}

Optical flow estimates pixel-wise motion and plays a fundamental role in dynamic scene understanding. Despite substantial advances in frame-based methods~\cite{dosovitskiy2015flownet,sun2018pwc,teed2020raft,xu2022gmflow}, high-speed motion remains challenging due to exposure-induced blur and limited temporal sampling. Neuromorphic sensors offer an alternative sensing paradigm. Unlike event cameras that respond to changes in logarithmic intensity~\cite{gallego2022eventvision}, spike cameras integrate incident light and emit binary spikes upon reaching a threshold. These integrate-and-fire mechanisms provide fine temporal resolution and encode scene intensity through firing activity, but individual spikes do not directly represent instantaneous image intensities~\cite{hu2022scflow,zhao2022spike2flow}.

\begin{figure}[!t]
    \centering
    \includegraphics[width=\linewidth]{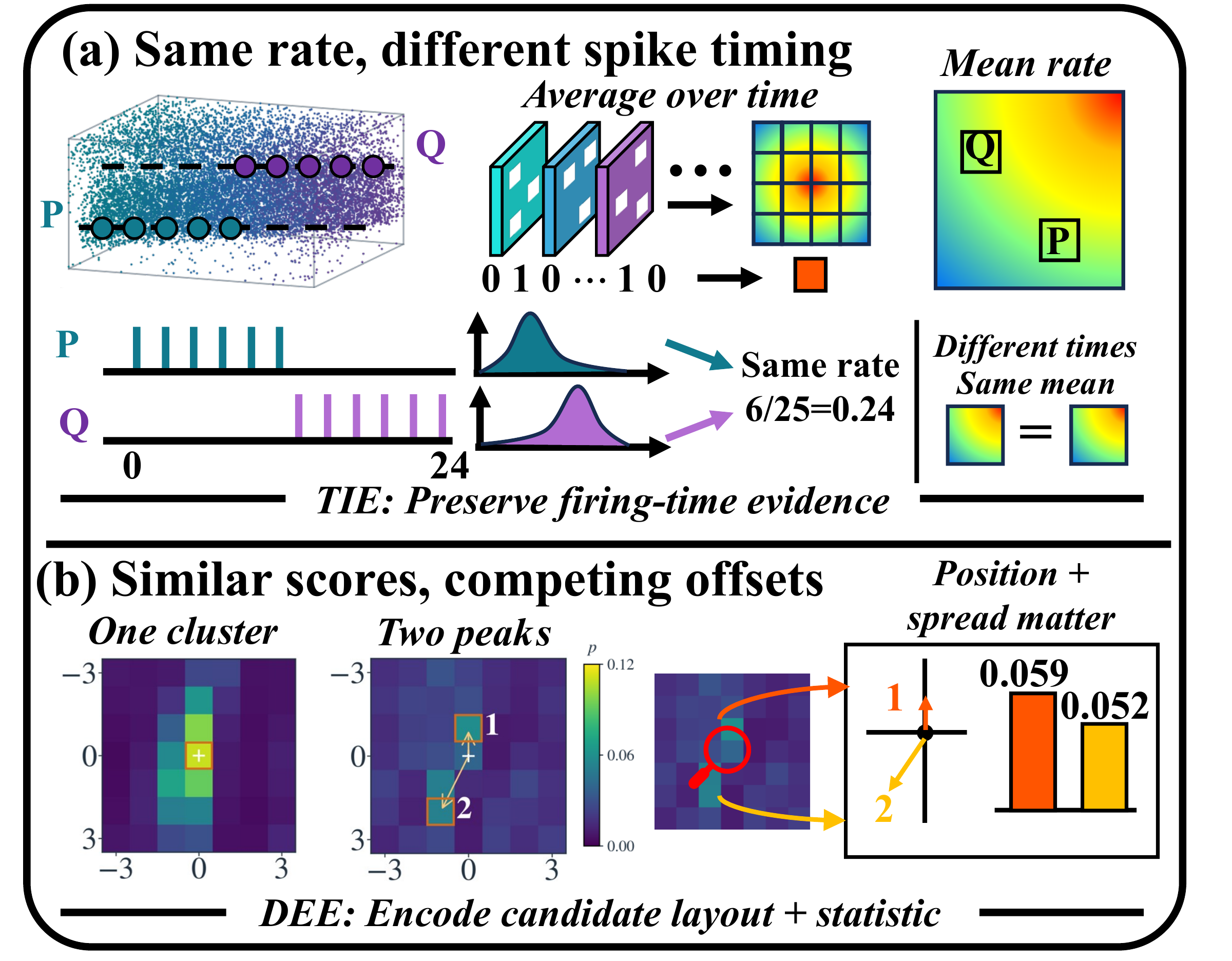}
    \caption{\textbf{Ambiguities in spike optical flow.}
    (a) Different firing sequences can yield the same firing rate.
    (b) Similar correlation scores can support distinct displacement hypotheses.}
    \label{fig:motivation}
\end{figure}

Estimating optical flow from spikes therefore requires motion representations adapted to their sensing characteristics. Existing methods have explored temporal sampling, firing-time intervals, spatiotemporal fusion, and motion aggregation~\cite{zhao2024hist,zhao2026spike2flowpp,zhai2025direction}. Related spike reconstruction studies have also investigated temporal aggregation across different scales~\cite{dong2025longshort}. However, temporal aggregation and spatial compression may weaken fine-grained firing-time cues. As illustrated in \cref{fig:motivation}(a), distinct spike patterns can produce identical firing rates, indicating that aggregated statistics cannot fully preserve their temporal configurations. Meanwhile, correlation-based flow estimators rely on learned matching features for correspondence reasoning~\cite{teed2020raft,huang2022flowformer}. Their local correlation distributions may contain competing candidates with similar scores but different displacement offsets, as shown in \cref{fig:motivation}(b). Although learned cost encoders can capture matching patterns~\cite{zhang2021separable,shi2023flowformerpp}, the spatial arrangement and distribution statistics of candidate displacements are not always explicitly represented. These observations highlight two complementary challenges: preserving informative spike timing and interpreting displacement evidence.

To address these challenges, we propose \textbf{TIDE-Flow}, a spike-based optical flow estimation framework that jointly exploits \emph{Temporal Integration Evidence (TIE)} and \emph{Displacement Evidence Encoding (DEE)}. First, TIE aims to explicitly preserve fine-grained firing information before spatial feature compression. It combines raw binary patterns, multi-scale firing rates, and a temporal first moment to retain complementary timing cues. Furthermore, boundary-aware inter-spike interval cues distinguish complete integration intervals from partially observed ones, providing additional information about the underlying firing process. To preserve local spatial organization, TIE rearranges neighboring spike measurements into separate channels before learned feature reduction. This enables the network to jointly exploit detailed firing configurations and complementary temporal statistics for distinct motion representation.

Second, DEE improves how local correspondence evidence is encoded during optical flow estimation. DEE arranges the sampled correlations on a two-dimensional grid of candidate offsets and encodes their local structure. It further introduces explicit displacement statistics, including candidate locations, spatial dispersion, and peak-related measurements, to characterize the concentration and competition of plausible correspondences. These learned and explicit representations provide complementary geometric evidence for subsequent flow refinement. TIDE-Flow integrates TIE-based temporal features and DEE-based displacement features into recurrent motion estimation, enabling both forms of evidence to guide iterative flow updates.

In summary, our main contributions are threefold.
\begin{itemize}
    \item We propose Temporal Integration Evidence (TIE) to preserve fine-grained spike timing through joint modeling of firing patterns, multi-scale statistics, and boundary-aware interval cues before spatial compression.

    \item We develop Displacement Evidence Encoding (DEE) to enhance correspondence reasoning by combining structured candidate-grid features with explicit statistics of displacement geometry and matching ambiguity.

    \item Extensive experiments on synthetic and real spike data demonstrate the effectiveness of TIDE-Flow, achieving competitive optical flow accuracy while striking a better balance between accuracy, model size, and computation.
\end{itemize}

\section{Related Work}

\subsection{Spike Camera Optical Flow}
Spike cameras encode accumulated illumination through binary firing streams, whose temporal characteristics and firing fluctuations have been studied in spike imaging~\cite{zheng2021stp,zhao2024fluctuations}. SCFlow~\cite{hu2022scflow} introduced motion-guided temporal sampling, while Spike2Flow~\cite{zhao2022spike2flow} exploited differential spike firing time (DSFT) and correlation matching. Subsequent methods investigated self-supervised flow estimation~\cite{chen2023usflow}, hierarchical spatiotemporal fusion~\cite{zhao2024hist}, dual-DSFT motion modeling~\cite{zhao2026spike2flowpp}, and direction-disassembled refinement~\cite{zhai2025direction}. Related reconstruction methods further explored temporal correlations and motion-aware aggregation~\cite{zhao2021spk2imgnet,dong2025csc,dong2025longshort}. TIDE-Flow organizes raw firing patterns, multi-scale statistics, and boundary-aware interval evidence before spatial compression, while incorporating structured displacement modeling for correspondence estimation.

\subsection{Event-based Optical Flow}
Unlike spike cameras, event cameras measure changes in logarithmic intensity without absolute intensity. Early methods developed event-specific representations and learning objectives~\cite{zhu2018evflownet,zhu2019unsup}. E-RAFT~\cite{gehrig2021eraft} introduced dense correlation and recurrent refinement, followed by advances in temporal aggregation~\cite{liu2023tma,ponghiran2023dense}, continuous-time estimation~\cite{wan2022dceiflow,gehrig2024dense,hu2026consistency}, and efficient flow modeling~\cite{luo2024eemflow}. Recent approaches further explored unified matching~\cite{zhang2025ematch}, dense temporal differences~\cite{liu2025edcflow}, and adaptive temporal correlation~\cite{xu2026bat}. Building on recurrent difference-correlation refinement, TIDE-Flow addresses the integrate-and-fire characteristics of spike cameras through fine-grained temporal evidence and structured displacement encoding.

\subsection{Correspondence and Matching-Cost Modeling}
Optical flow estimation has progressed from pyramid-based cost volumes~\cite{sun2018pwc} to recurrent all-pairs matching~\cite{teed2020raft,wang2024searaft}. Subsequent methods improved correspondence reasoning through global aggregation~\cite{jiang2021gma}, neighborhood affinity~\cite{luo2022kpaflow}, structured cost modeling~\cite{zhang2021separable}, and transformer-based matching~\cite{xu2022gmflow,huang2022flowformer,sui2022craft,shi2023flowformerpp}. Recent studies also explored efficient correlation sampling~\cite{briedis2026sampling} and temporal memory~\cite{dong2024memflow}. DEE combines learned patterns on the sampled candidate grid with geometric moments and peak statistics to characterize competing correspondences.

\section{Proposed Method}

\subsection{Overview}

\begin{figure*}[!t]
    \centering
    \includegraphics[width=\linewidth]{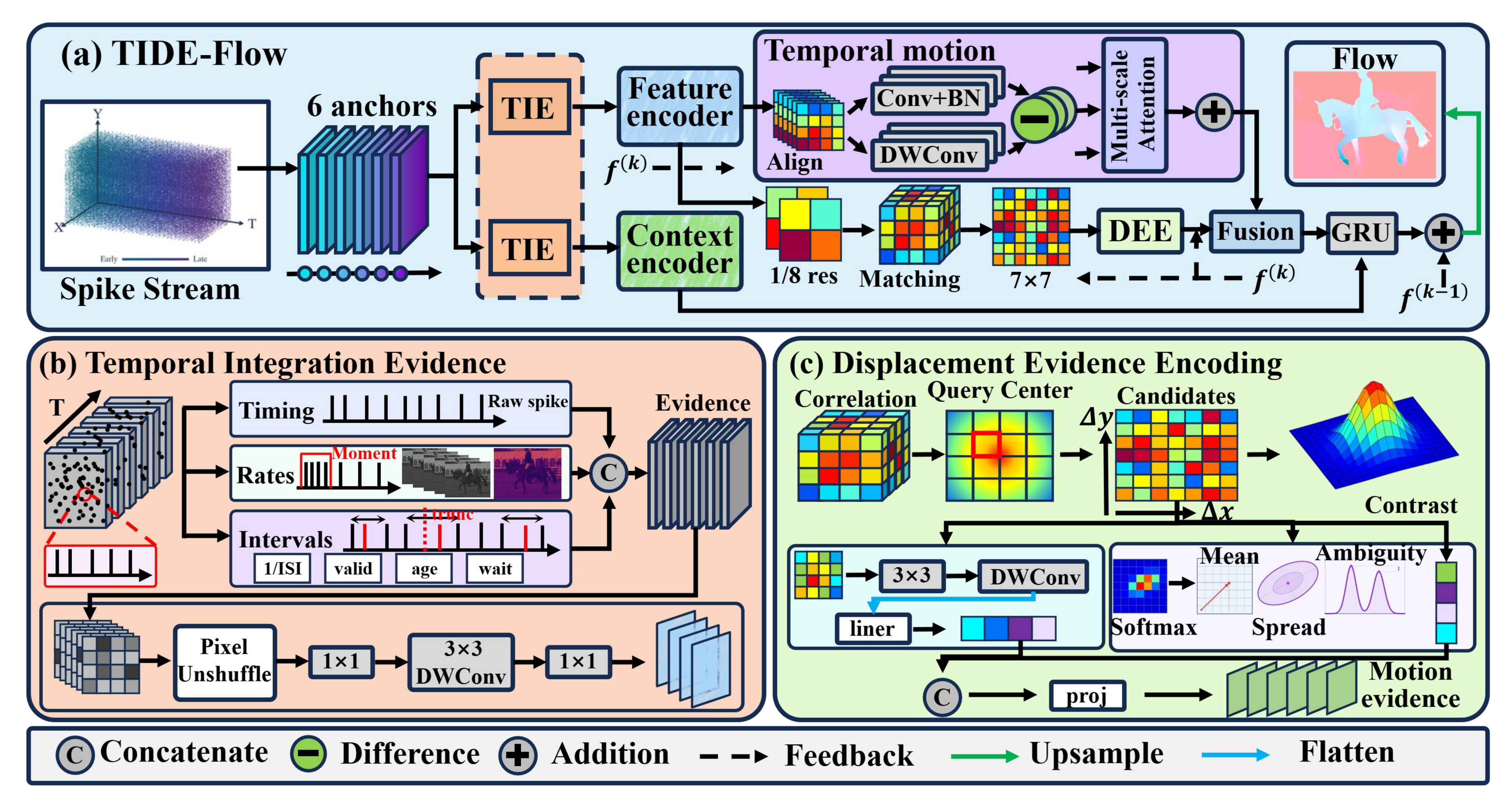}
    \caption{\textbf{TIDE-Flow architecture.}
    (a) Each flow update combines temporal motion evidence with local displacement evidence.
    (b) TIE preserves firing patterns and inter-spike interval cues before spatial compression.
    (c) DEE combines candidate-grid features with explicit displacement statistics.}
    \label{fig:framework}
\end{figure*}

We propose \textbf{TIDE-Flow}, a spike-based optical flow estimation framework that integrates Temporal Integration Evidence (TIE) and Displacement Evidence Encoding (DEE) for motion estimation from binary spike streams. As illustrated in \cref{fig:framework}(a), TIE organizes complementary temporal measurements before feature compression, while DEE encodes the geometric structure and distribution of candidate correspondences after feature matching.

Given a spike stream $\bm{S}\in\left\{0,1\right\}^{T\times H\times W}$, our objective is to estimate dense optical flow $\bm{f}\in\mathbb{R}^{2\times H\times W}$ between two reference times. We construct six temporal anchors, each containing a local window of 25 spike frames. After TIE encoding, the resulting representations are processed by feature and context encoders to support two complementary pathways: temporal motion modeling at $1/4$ resolution and correspondence matching at $1/8$ resolution.

We adopt a recurrent difference-correlation refinement scheme. The temporal pathway extracts motion cues from aligned feature differences, while the matching pathway retrieves flow-guided correlation costs and processes them through DEE. Their outputs are fused to update the optical flow iteratively, with each prediction guiding the subsequent temporal alignment and correspondence lookup.

\subsection{Temporal Integration Evidence Encoding}

As shown in \cref{fig:framework}(b), TIE effectively combines fine-grained spike timing with complementary integration statistics and boundary-aware interval measurements. This allows the network to retain distinct temporal configurations that may be obscured by firing-rate aggregation.

\vspace{-10pt}
\paragraph{Temporal Evidence Construction.}
For the $j$-th temporal anchor, let $\bm{S}_j\in\left\{0,1\right\}^{L\times H\times W}$ denote the corresponding local spike sequence, where $L=25$. We retain the original binary observations and construct two additional types of complementary temporal evidence.

First, multi-scale firing rates and a temporal first moment characterize spike activity over different temporal ranges:

\begin{equation}
\begin{aligned}
    r_j^{(\ell)}\left(\bm{x}\right) &= \frac{1}{\left|\mathcal{W}_{\ell}\right|}\sum_{t\in\mathcal{W}_{\ell}}S_j\left(t,\bm{x}\right), \\
    m_j\left(\bm{x}\right) &= \frac{\sum_{t=0}^{L-1}\hat{t} S_j\left(t,\bm{x}\right)}{\sum_{t=0}^{L-1}S_j\left(t,\bm{x}\right)+\epsilon },
\end{aligned}
\end{equation}
where $\mathcal{W}_{\ell}$ denotes a centered temporal window with $\ell\in\left\{3,7,13,25\right\}$, $\hat{t}$ is the normalized temporal coordinate, and $\epsilon$ ensures numerical stability. The firing rates describe activity at multiple temporal scales, while the first moment captures its temporal distribution.

Second, we introduce boundary-aware inter-spike interval evidence. We use five probe positions within each anchor window. At each probe $q$, the nearest preceding and succeeding spike times are identified as $t^-_{j,q}$ and $t^+_{j,q}$, respectively. Four complementary measurements are constructed:

\begin{equation}
\begin{aligned}
    \bm{b}_{j,q}(\bm{x}) &= [\nu_{j,q},v_{j,q},a_{j,q},w_{j,q}],\\
    v_{j,q} &= \mathbb{I}\left[
    t^-_{j,q}\neq\varnothing \land
    t^+_{j,q}\neq\varnothing
    \right],
\end{aligned}
\end{equation}
where $\nu_{j,q}$ is the reciprocal inter-spike interval, $v_{j,q}$ indicates whether both interval endpoints are observed, and $a_{j,q}$ and $w_{j,q}$ denote the elapsed and waiting times relative to the probe. For a fully observed interval, $\nu_{j,q}=\left(t^+_{j,q}-t^-_{j,q}+\epsilon\right)^{-1}$. When an endpoint is missing, $v_{j,q}=0$ flags a boundary-truncated interval.

The temporal evidence is concatenated as:
\begin{equation}
\begin{aligned}
    \bm{E}_j &= \operatorname{Concat} \left( \bm{S}_j,\bm{r}_j, \bm{m}_j,\bm{b}_j \right) \in\mathbb{R}^{50\times H\times W},
\end{aligned}
\end{equation}
where $\bm{r}_j$ contains four firing-rate maps and $\bm{b}_j$ contains four interval measurements at each of five probe positions. Together with the 25 raw spike channels and one temporal-moment channel, TIE produces a 50-channel representation.

\vspace{-10pt}
\paragraph{Spatially Organized Evidence Encoding.}
To preserve local sampling positions before learned spatial reduction, we rearrange each $2\times2$ neighborhood into separate channels using pixel unshuffling:
\begin{equation}
\begin{aligned}
    \widetilde{\bm{E}}_j &= \operatorname{PixelUnshuffle}_{2} \left(\bm{E}_j\right) \in\mathbb{R}^{200\times H/2\times W/2}, \\
    \bm{Z}_j &= \operatorname{PW}_2 \left( \operatorname{DWConv}_{3\times3} \left( \operatorname{PW}_1 \left(\widetilde{\bm{E}}_j\right) \right) \right),
\end{aligned}
\end{equation}
where $\operatorname{PW}$ and $\operatorname{DWConv}$ denote pointwise and depthwise convolutions, respectively, with intermediate normalization and activation omitted for brevity. Unlike direct spatial averaging, pixel unshuffling preserves the distinct positions of neighboring spike measurements through channel rearrangement. The subsequent convolutions integrate temporal and spatial evidence before the resulting features are passed to the motion and matching encoders.

\subsection{Displacement Evidence Encoding}

DEE models competing correspondences by combining learned candidate-grid features with explicit displacement statistics, as illustrated in \cref{fig:framework}(c).

\vspace{-10pt}
\paragraph{Flow-Guided Correlation Lookup.}
Let $\bm{F}_0^m$ and $\bm{F}_1^m$ denote matching features at the two reference times, extracted at $1/8$ resolution. For the full model, we apply feature normalization before forming the all-pairs correlation volume:
\begin{equation}
\begin{aligned}
    \overline{\bm{F}}_i^m\left(\bm{x}\right) &= \frac{\bm{F}_i^m\left(\bm{x}\right)} {\left\|\bm{F}_i^m\left(\bm{x}\right)\right\|_2+\epsilon}, \quad i\in\left\{0,1\right\}, \\
    \mathcal{C}\left(\bm{x},\bm{y}\right) &= \kappa\left\langle \overline{\bm{F}}_0^m\left(\bm{x}\right), \overline{\bm{F}}_1^m\left(\bm{y}\right) \right\rangle,
\end{aligned}
\end{equation}
where $\kappa$ scales the correlation responses. This channel-wise normalization reduces the dependence of matching scores on feature magnitude. It is independently configurable: the \emph{Norm.} ablation denotes whether matching features are normalized before correlation, while the unnormalized variant directly correlates $\bm{F}_0^m$ and $\bm{F}_1^m$. Importantly, feature normalization is distinct from the candidate-score centering within DEE described below. The all-pairs correlation volume is computed once and reused across refinement iterations.

At iteration $k$, the current estimate is mapped from the $1/4$ update grid to the $1/8$ matching grid. We then sample the correlation volume around the predicted correspondence:

\begin{equation}
\begin{aligned}
    \bm y_8^{(k)} \left(\bm x_8\right)
        &= \bm x_8 + \widehat{\bm f}_8^{(k-1)} \left(\bm x_8\right),\\
    Q_8^{(k)} \left(\bm x_8,\bm\delta\right)
        &= \operatorname{Bilinear} \left(
        \mathcal C \left(\bm x_8,\cdot\right),
        \bm y_8^{(k)} \left(\bm x_8\right)+\bm\delta\right),\\
    Q_4^{(k)} \left(\cdot,\bm\delta\right)
        &= \operatorname{Up}_{2} \left(Q_8^{(k)} \left(\cdot,\bm\delta\right)\right),
\end{aligned}
\end{equation}
where $\bm x_8$ is a matching-grid position, and $\widehat{\bm f}_8^{(k-1)}$ is the previous flow resized and expressed in matching-grid pixels. The offset $\bm\delta$ belongs to $\mathcal D=\left\{-3,\ldots,3\right\}^{2}$ in the default model. Upsampling acts on the image positions of each of the 49 cost channels before DEE. The channels retain their respective candidate offsets.

\vspace{-10pt}
\paragraph{Structured Candidate-Grid Encoding.}
For each update-grid position, we reshape its 49 matching scores into a $7\times7$ displacement grid. To emphasize relative matching responses, we subtract their mean within each candidate grid, independently of the matching-feature normalization:
\begin{equation}
\begin{aligned}
    \widetilde{Q}^{(k)}_{\bm{\delta}} &= Q^{(k)}_{\bm{\delta}} - \frac{1}{49} \sum_{\bm{\delta}'} Q^{(k)}_{\bm{\delta}'}, \\
    \bm{g}_{\mathrm{grid}}^{(k)} &= \operatorname{Linear} \left( \operatorname{Flatten} \left( \operatorname{DWConvs} \left(\widetilde{\bm{Q}}^{(k)}  \right) \right) \right),
\end{aligned}
\end{equation}
where $Q^{(k)}_{\bm{\delta}}$ denotes a score from $\bm{Q}_4^{(k)}$ at a given update-grid location. The convolutions operate over the candidate displacement grid, allowing nearby displacement hypotheses to be modeled jointly. This produces a learned descriptor $\bm{g}_{\mathrm{grid}}^{(k)}$ that captures structured correlation patterns.

\vspace{-10pt}
\paragraph{Explicit Displacement Statistics.}
To complement the learned representation, we normalize the candidate scores into a distribution:
\begin{equation}
\begin{aligned}
    p_{\bm{\delta}}^{(k)} &= \frac{ \exp \left(Q_{\bm{\delta}}^{(k)}\right) }{ \sum_{\bm{\delta}'} \exp \left(Q_{\bm{\delta}'}^{(k)}\right) },
\end{aligned}
\end{equation}
where each candidate has a normalized two-dimensional offset $\bm{u}_{\bm{\delta}}$.
We compute the displacement mean and covariance:
\begin{equation}
\begin{aligned}
    \bm{\mu}^{(k)} &= \sum_{\bm{\delta}} p_{\bm{\delta}}^{(k)}
    \bm{u}_{\bm{\delta}}, \\
    \bm{\Sigma}^{(k)} &= \sum_{\bm{\delta}} p_{\bm{\delta}}^{(k)} \left( \bm{u}_{\bm{\delta}} -\bm{\mu}^{(k)} \right) \left( \bm{u}_{\bm{\delta}} -\bm{\mu}^{(k)} \right)^\top.
\end{aligned}
\end{equation}

The mean describes the center of the candidate distribution, while the covariance characterizes its spatial spread and directional variation.

We additionally measure distribution entropy, peak probability, the gap between the two largest probabilities, and peak-to-mean score contrast. These quantities form a nine-dimensional descriptor:
\begin{equation}
\begin{aligned}
    \bm{g}_{\mathrm{stat}}^{(k)} &= \left[\bm{\mu}^{(k)}, \operatorname{vech} \left(\bm{\Sigma}^{(k)}\right), H^{(k)},p_{(1)}^{(k)}, \Delta p^{(k)},\Delta c^{(k)} \right],
\end{aligned}
\end{equation}
where $\operatorname{vech}\left(\bm{\Sigma}\right)$ contains its three unique covariance entries, $H=-\sum_{\bm\delta\in\mathcal D}p_{\bm\delta}\log \left(p_{\bm\delta}\right)/\log\left|\mathcal D\right|$ denotes normalized entropy, $p_{(1)}$ is the largest candidate probability, $\Delta p=p_{(1)}-p_{(2)}$ is the top-two probability gap, and $\Delta c=\max_{\bm{\delta}}Q_{\bm{\delta}}-\operatorname{Mean}_{\bm{\delta}}Q_{\bm{\delta}}$ measures peak-to-mean contrast. These statistics explicitly summarize displacement geometry and the concentration of competing matches.

Finally, the learned candidate-grid features and explicit statistics are fused:
\begin{equation}
\begin{aligned}
    \bm{G}^{(k)} &= \operatorname{Proj} \left( \left[ \bm{g}_{\mathrm{grid}}^{(k)}, \bm{g}_{\mathrm{stat}}^{(k)} \right] \right),
\end{aligned}
\end{equation}
yielding displacement evidence $\bm{G}^{(k)}$ for refinement.

\subsection{Evidence-Guided Flow Refinement}

\paragraph{Temporal Motion Modeling.}
The temporal branch operates on six TIE-derived anchor features at $1/4$ resolution. Following temporal difference-based motion modeling~\cite{liu2025edcflow}, features are aligned using the current flow estimate and their relative temporal positions. Differences are extracted over anchor strides of $1$, $2$, and $5$, then processed by convolutional transformations and multi-scale attention to obtain temporal motion evidence $\bm{D}^{(k)}$.

\vspace{-10pt}
\paragraph{Evidence Fusion and Flow Update.}
At each iteration, temporal motion evidence $\bm{D}^{(k)}$ and displacement evidence $\bm{G}^{(k)}$ are fused with context features $\bm{c}$ and the current flow estimate. A convolutional GRU then updates its hidden state and predicts a residual flow:
\begin{equation}
\begin{aligned}
    \bm{z}^{(k)} &= \operatorname{Fuse} \left( \bm{D}^{(k)}, \bm{G}^{(k)}, \bm{c}, \bm{f}_4^{(k-1)} \right), \\
    \bm{h}^{(k)} &= \operatorname{ConvGRU} \left( \bm{h}^{(k-1)}, \bm{z}^{(k)} \right), \\
    \bm{f}_4^{(k)} &= \bm{f}_4^{(k-1)} + \operatorname{FlowHead} \left(\bm{h}^{(k)}\right).
\end{aligned}
\end{equation}

The updated flow realigns the temporal features and recenters the correlation lookup at the next iteration.

Starting from zero flow, TIDE-Flow performs $K=6$ recurrent updates. Each intermediate flow is converted to full resolution through learned convex upsampling. Following~\cite{teed2020raft,liu2025edcflow}, the training objective is:
\begin{equation}
\begin{aligned}
    \mathcal{L}_{\mathrm{flow}} &= \sum_{k=1}^{K} \gamma^{K-k} \frac{1}{\left|\Omega\right|} \sum_{\bm{x}\in\Omega} \left\| \widehat{\bm{f}}^{(k)}\left(\bm{x}\right) -\bm{f}_{\mathrm{GT}}\left(\bm{x}\right) \right\|_1,
\end{aligned}
\end{equation}
where $\widehat{\bm{f}}^{(k)}$ is the upsampled prediction, $\Omega$ denotes valid pixels, and $\gamma=0.8$ controls the contribution of intermediate predictions. The final iteration produces the full-resolution optical flow estimate.

\begin{table*}[!t]
    \centering
    \footnotesize
    \setlength{\tabcolsep}{3pt}
    \caption{\textbf{Optical flow accuracy on PHM.}
    Entries report AEPE (pixels) / PO (\%).
    Boldface and underlining mark the best and second-best distinct values,
    respectively, for each metric, scene, and temporal interval.}
    \begin{tabularx}{\linewidth}{@{}c|l|*{9}C|c@{}}
    \toprule[1pt]
    & {Method} & {Ball} & {Cook} & {Dice} & {Doll} & {Fan} & {Hand} & {Jump} & {Poker} & {Top} & {Average} \\
    \midrule
    \multirow{14}{*}{\rotatebox[origin=c]{90}{$\Delta t=10$}} 
    & RAFT~\cite{teed2020raft} & 0.46 / 12.5 & 1.32 / 43.7 & 0.95 / 29.3 & 0.24 / 6.7 & 0.28 / 12.7 & 1.11 / 45.1 & 0.11 / 3.0 & 0.67 / 37.1 & 2.19 / 19.7 & 0.813 / 23.30 \\
    & Flow1D~\cite{xu2021flow1d} & 0.79 / 51.4 & 1.28 / 50.8 & 1.15 / 47.9 & 0.27 / 6.3 & 0.28 / 11.0 & 1.86 / 83.1 & 0.13 / 3.4 & 0.85 / 50.1 & 2.19 / 17.7 & 0.979 / 35.76 \\
    & FlowFormer~\cite{huang2022flowformer} & 0.52 / 13.5 & 1.48 / 58.7 & 0.98 / 31.0 & 0.25 / 6.7 & 0.29 / 11.5 & 1.82 / 84.5 & 0.14 / 3.6 & 0.94 / 54.9 & 2.22 / 19.5 & 0.959 / 31.54 \\
    & SCFlow~\cite{hu2022scflow} & 0.51 / 20.3 & 1.34 / 38.6 & 1.10 / 30.7 & 0.22 / 5.6 & 0.24 / 10.7 & 1.30 / 57.3 & 0.11 / 3.0 & 0.80 / 41.1 & 2.14 / 17.7 & 0.863 / 25.00 \\
    & Spike2Flow~\cite{zhao2022spike2flow} & 0.39 / 12.3 & 1.61 / 46.5 & \underline{0.66} / \underline{20.4} & 0.25 / 7.4 & 0.27 / 10.9 & 1.60 / 62.4 & 0.14 / 4.3 & 0.65 / 26.2 & 2.18 / 15.7 & 0.861 / 22.89 \\
    & CRAFT~\cite{sui2022craft} & 0.61 / 15.0 & 1.28 / 43.5 & 0.93 / 27.6 & 0.19 / 5.0 & 0.25 / 10.2 & 1.67 / 73.3 & 0.10 / 2.6 & 0.56 / 23.1 & 2.15 / 15.1 & 0.860 / 23.94 \\
    & KPA-Flow~\cite{luo2022kpaflow} & 0.47 / 14.9 & 1.41 / 45.9 & 0.87 / 29.9 & 0.27 / 7.1 & 0.29 / 12.7 & 1.19 / 47.7 & 0.12 / 3.0 & 0.65 / 36.6 & 2.19 / 19.4 & 0.827 / 24.12 \\
    & EEMFlow+~\cite{luo2024eemflow} & 0.37 / 11.3 & 1.38 / 41.3 & 1.10 / 30.1 & 0.21 / 7.0 & 0.23 / 11.1 & 1.56 / 47.2 & 0.11 / 3.8 & 0.63 / 23.1 & 2.19 / 19.8 & 0.864 / 21.64 \\
    & HiST-SFlow~\cite{zhao2024hist} & \textbf{0.28} / \underline{7.80} & \underline{0.80} / \underline{27.4} & 0.85 / 23.3 & 0.20 / 5.6 & 0.27 / 12.8 & \textbf{0.64} / \textbf{21.7} & 0.08 / 2.5 & 0.53 / 23.9 & \underline{2.11} / 14.8 & 0.640 / 15.54 \\
    & EMatch~\cite{zhang2025ematch} & 0.46 / 10.0 & 1.32 / 60.9 & 0.94 / 37.3 & 0.45 / 23.4 & 0.36 / 8.80 & 1.24 / 61.7 & 0.28 / 5.2 & 0.58 / \underline{17.7} & 2.33 / 28.5 & 0.884 / 28.16 \\
    & EDCFlow~\cite{liu2025edcflow} & 0.57 / 9.90 & 0.99 / 35.7 & 0.76 / 20.5 & \underline{0.16} / \underline{4.5} & 0.20 / 9.50 & 1.26 / 50.0 & \underline{0.06} / 2.0 & 0.52 / 22.3 & \underline{2.11} / 14.7 & 0.738 / 18.79 \\
    & SIV~\cite{zhang2026siv} & 0.34 / 8.20 & \textbf{0.65} / \textbf{26.3} & 0.94 / 24.6 & \underline{0.16} / 4.6 & 0.19 / \underline{8.50} & 0.70 / 25.5 & \underline{0.06} / \underline{1.9} & 0.49 / 26.1 & \textbf{2.10} / 15.2 & 0.625 / 15.65 \\
    & Spike2Flow++~\cite{zhao2026spike2flowpp} & 0.31 / 8.50 & 0.97 / 27.9 & 0.70 / 21.5 & 0.17 / 5.1 & \underline{0.18} / 9.30 & \underline{0.69} / \underline{23.6} & 0.07 / 2.4 & \textbf{0.37} / \textbf{14.3} & \underline{2.11} / \underline{14.4} & \underline{0.620} / \underline{14.12} \\
     \rowcolor{gray!15}
    & {TIDE-Flow} (Ours) & \underline{0.29} / \textbf{7.40} & 0.91 / 30.0 & \textbf{0.51} / \textbf{14.7} & \textbf{0.15} / \textbf{3.9} & \textbf{0.16} / \textbf{6.90} & 0.70 / 26.2 & \textbf{0.05} / \textbf{1.4} & \underline{0.39} / \textbf{14.3} & \textbf{2.10} / \textbf{14.0} & \textbf{0.584} / \textbf{13.21} \\
    \midrule[1pt]
    \multirow{14}{*}{\rotatebox[origin=c]{90}{$\Delta t=20$}}
    & RAFT~\cite{teed2020raft} & 0.78 / 18.6 & 2.75 / 54.4 & 1.57 / 30.1 & 0.43 / 9.30 & 0.50 / 14.6 & 2.81 / 59.9 & 0.21 / 5.80 & 1.31 / 46.7 & 4.30 / 21.2 & 1.628 / 28.94 \\
    & Flow1D~\cite{xu2021flow1d} & 1.19 / 51.6 & 4.52 / 96.3 & 1.58 / 50.7 & 0.78 / 53.3 & 1.01 / 82.1 & 6.65 / 99.2 & 0.72 / 73.1 & 1.39 / 52.3 & 4.75 / 79.7 & 2.510 / 70.90 \\
    & FlowFormer~\cite{huang2022flowformer} & 0.91 / 13.8 & 4.41 / 96.3 & 1.40 / 32.6 & 0.80 / 54.8 & 1.03 / 90.0 & 6.54 / 99.3 & 0.74 / 75.8 & 1.47 / 57.4 & 4.59 / 61.9 & 2.432 / 64.67 \\
    & SCFlow~\cite{hu2022scflow} & 0.94 / 27.1 & 3.00 / 50.6 & 1.72 / 33.2 & 0.41 / 8.10 & 0.46 / 13.6 & 3.71 / 71.3 & 0.19 / 5.90 & 1.57 / 53.7 & 4.25 / 18.9 & 1.804 / 31.37 \\
    & CRAFT~\cite{sui2022craft} & 1.16 / 85.5 & 2.68 / 61.0 & 1.99 / 46.8 & 0.39 / 7.80 & 0.48 / 12.5 & 3.53 / 87.1 & 0.20 / \textbf{3.60} & 1.23 / 38.9 & 4.31 / 22.0 & 1.775 / 40.57 \\
    & Spike2Flow~\cite{zhao2022spike2flow} & 0.80 / 16.8 & 3.13 / 65.2 & \underline{1.31} / 29.3 & 0.48 / 11.8 & 0.53 / 15.3 & 3.18 / 73.1 & 0.29 / 7.80 & 1.32 / 39.9 & 4.31 / 22.1 & 1.705 / 31.28 \\
    & KPA-Flow~\cite{luo2022kpaflow} & 0.80 / 20.9 & 2.93 / 55.6 & 1.48 / 31.4 & 0.45 / 9.60 & 0.52 / 14.5 & 2.86 / 62.5 & 0.22 / 5.60 & 1.31 / 48.4 & 4.28 / 19.7 & 1.649 / 29.81 \\
    & EEMFlow+~\cite{luo2024eemflow} & 0.72 / 12.5 & 2.68 / 46.7 & 2.45 / 34.3 & 0.41 / 9.50 & 0.45 / 13.9 & 3.02 / 56.1 & 0.21 / 5.30 & 1.24 / 34.3 & 4.34 / 30.4 & 1.725 / 26.99 \\
    & HiST-SFlow~\cite{zhao2024hist} & \textbf{0.55} / \underline{8.80} & 2.04 / \textbf{33.6} & 1.64 / 26.3 & 0.38 / 7.20 & 0.51 / 13.9 & 2.00 / 34.7 & 0.17 / 5.00 & 1.28 / 33.1 & \underline{4.18} / \underline{15.1} & 1.417 / 19.73 \\
    & EMatch~\cite{zhang2025ematch} & 0.78 / 12.7 & 2.42 / 63.3 & 1.69 / 57.1 & 0.76 / 51.6 & 0.66 / 49.3 & 2.24 / 55.5 & 0.51 / 37.3 & 1.05 / 29.4 & 4.55 / 49.8 & 1.628 / 45.12 \\
    & EDCFlow~\cite{liu2025edcflow} & 1.13 / 89.4 & 2.41 / 57.0 & 1.57 / 54.1 & 0.34 / 7.00 & 0.41 / 12.0 & 3.07 / 74.3 & 0.16 / 4.80 & 1.20 / 47.4 & 4.23 / 19.5 & 1.613 / 40.61 \\
    & SIV~\cite{zhang2026siv} & 0.63 / 9.50 & \textbf{1.37} / \underline{34.0} & 1.94 / 28.7 & \underline{0.31} / \underline{6.40} & \underline{0.35} / \underline{11.2} & \underline{1.66} / 36.2 & \underline{0.13} / 5.00 & 0.95 / 35.9 & \underline{4.18} / 16.3 & 1.282 / 20.37 \\
    & Spike2Flow++~\cite{zhao2026spike2flowpp} & 0.62 / 9.50 & \underline{1.91} / 34.4 & 1.41 / \underline{26.1} & 0.32 / 6.80 & \underline{0.35} / 12.6 & \textbf{1.41} / \textbf{32.5} & 0.15 / 5.00 & \textbf{0.76} / \textbf{20.5} & 4.19 / 15.7 & \textbf{1.236} / \underline{18.11} \\
     \rowcolor{gray!15}
    & {TIDE-Flow} (Ours) & \underline{0.58} / \textbf{8.10} & 2.00 / 35.4 & \textbf{1.01} / \textbf{18.4} & \textbf{0.30} / \textbf{5.60} & \textbf{0.31} / \textbf{9.20} & 1.82 / \underline{33.5} & \textbf{0.12} / \underline{4.50} & \underline{0.88} / \underline{21.9} & \textbf{4.16} / \textbf{14.3} & \underline{1.241} / \textbf{16.75} \\
    \bottomrule[1pt]
    \end{tabularx}
    \label{tab:phm_comparison}
\end{table*}

\section{Experimental Results}

\paragraph{Dataset.}
Following prior spike-based optical flow studies~\cite{zhao2024hist,zhao2026spike2flowpp}, we train on SPIFT and evaluate on PHM, both introduced by SCFlow~\cite{hu2022scflow}. SPIFT contains 100 training and 10 validation sequences (50K and 5K samples), while PHM comprises multiple photorealistic scenes with challenging motions, including rotation, collision, and camera movement. Both datasets provide simulated spike streams and dense ground-truth flow. Following the established evaluation protocol~\cite{hu2022scflow,zhao2022spike2flow}, we evaluate at $\Delta t\in\left\{10,20\right\}$ to assess different motion magnitudes. We further conduct qualitative evaluations on real high-speed spike recordings.

\vspace{-10pt}
\paragraph{Evaluation Metrics.}
We evaluate optical flow using Average End-Point Error (AEPE)~\cite{baker2011database} and Percentage of Outliers (PO\%)~\cite{zhao2022spike2flow,zhao2024hist}. AEPE measures the mean endpoint error, while PO\% measures the fraction of pixels with errors exceeding both 0.5 pixels and 5\% of the ground-truth flow magnitude~\cite{zhao2022spike2flow}. Lower is better for both metrics.

\vspace{-10pt}
\paragraph{Implementation Details.}
We train TIDE-Flow on SPIFT using two 25-frame spike windows to construct six temporal anchors. Training employs random $320\times448$ crops and horizontal/vertical flips, with AdamW (weight decay $10^{-4}$, batch size 3). We train for 100K iterations using OneCycle scheduling with a peak learning rate of $2\times10^{-4}$. Following~\cite{liu2025edcflow}, we perform six recurrent updates with sequence-weighted $\ell_1$ supervision ($\gamma=0.8$).

\begin{table}[t]
    \centering
    \footnotesize
    \setlength{\tabcolsep}{2pt}
    \caption{Quantitative comparison of accuracy and inference cost.}
    \begin{tabularx}{\linewidth}{@{}l|ccc|cc@{}}
    \toprule[1.1pt]
    Method  & Params $\downarrow$ & MACs $\downarrow$ & Time $\downarrow$ & $\Delta t=10$ & $\Delta t=20$ \\
    \midrule
    HiST-SFlow  & \underline{7.904} & 1291.8 & 374.1 & 0.640 / 15.54 & 1.417 / 19.73 \\
    SIV & 8.068 & 2697.8 & 428.4 & 0.625 / 15.65 & 1.282 / 20.37 \\
    Spike2Flow++ & 8.650 & \underline{1216.6} & \underline{226.7} & \underline{0.620} / \underline{14.12} & \textbf{1.236} / \underline{18.11} \\
    \rowcolor{gray!20}
    {TIDE-Flow} & \textbf{2.519} & \textbf{692.0} & \textbf{168.7} & \textbf{0.584} / \textbf{13.21} & \underline{1.241} / \textbf{16.75} \\
    \bottomrule[1.1pt]
    \end{tabularx}
    \label{tab:phm_efficiency_paired}
\end{table}

\subsection{Comparative Experiments}

We compare TIDE-Flow with representative methods on the PHM~\cite{hu2022scflow} and real spike recordings. The compared methods are grouped into three categories: (1) spike-based methods, including SCFlow~\cite{hu2022scflow}, HiST-SFlow~\cite{zhao2024hist}, Spike2Flow~\cite{zhao2022spike2flow}, Spike2Flow++~\cite{zhao2026spike2flowpp}, and SIV~\cite{zhang2026siv}; (2) event-based methods, including EMatch~\cite{zhang2025ematch}, EEMFlow+~\cite{luo2024eemflow}, and EDCFlow~\cite{liu2025edcflow}; and (3) conventional frame-based optical flow methods, including RAFT~\cite{teed2020raft}, Flow1D~\cite{xu2021flow1d}, KPA-Flow~\cite{luo2022kpaflow}, CRAFT~\cite{sui2022craft}, and FlowFormer~\cite{huang2022flowformer}. For fair comparison, all methods are evaluated in the same experimental setting under identical evaluation protocols.

\vspace{-10pt}
\paragraph{Quantitative Comparisons.}
\cref{tab:phm_comparison} presents the quantitative results on the PHM benchmark~\cite{hu2022scflow} under two temporal intervals, $\Delta t=10$ and $\Delta t=20$. For $\Delta t=10$, TIDE-Flow achieves the best average AEPE of 0.584 and PO\% of 13.21, reducing the errors by 5.8\% and 6.4\%, respectively, compared with Spike2Flow++. Notably, our method achieves leading performance on several challenging scenes, including Dice, Doll, Fan, and Jump, demonstrating its effectiveness across diverse motion patterns. When the temporal interval increases to $\Delta t=20$, TIDE-Flow maintains competitive accuracy, achieving an AEPE of 1.241 and the lowest PO\% of 16.75. Compared with Spike2Flow++, it achieves a comparable AEPE while reducing PO\% from 18.11 to 16.75. These results indicate that TIDE-Flow provides accurate motion estimation while effectively reducing large displacement errors, particularly under increased temporal separation.

\vspace{-10pt}
\paragraph{Computational Efficiency.}
\cref{tab:phm_efficiency_paired} further compares the model complexity and inference efficiency at a resolution of $500\times800$. TIDE-Flow achieves the lowest computational cost among the compared methods, requiring only 2.519M parameters, 692.0G MACs, and 168.7 ms per inference. Compared with Spike2Flow++, our method reduces the parameter count by 70.9\%, MACs by 43.1\%, and inference time by 25.6\%, while achieving better average PO\% under both temporal intervals. These results demonstrate that TIDE-Flow achieves a favorable balance between optical flow accuracy and computational efficiency.

\begin{figure*}[!t]
    \centering
    \setlength{\tabcolsep}{1pt}
    \begin{tabular}{@{}ccccccccc@{}}
    \scriptsize Scene & \scriptsize EMatch & \scriptsize EEMFlow+ & \scriptsize Spike2Flow & \scriptsize Spike2Flow\texttt{++} & \scriptsize SIV & \scriptsize HiST-SFlow & \scriptsize \textbf{TIDE-Flow (Ours)} & \scriptsize Ground truth \\
    \includegraphics[width=0.1041\linewidth]{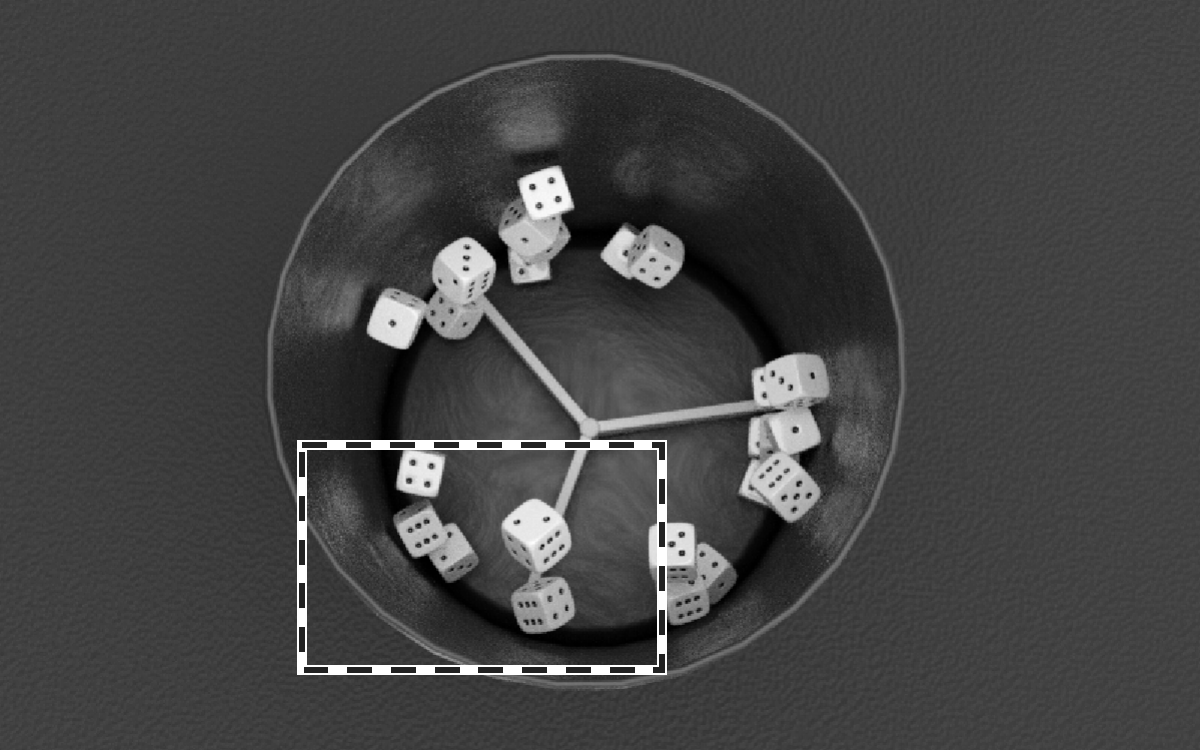} & \includegraphics[width=0.1041\linewidth]{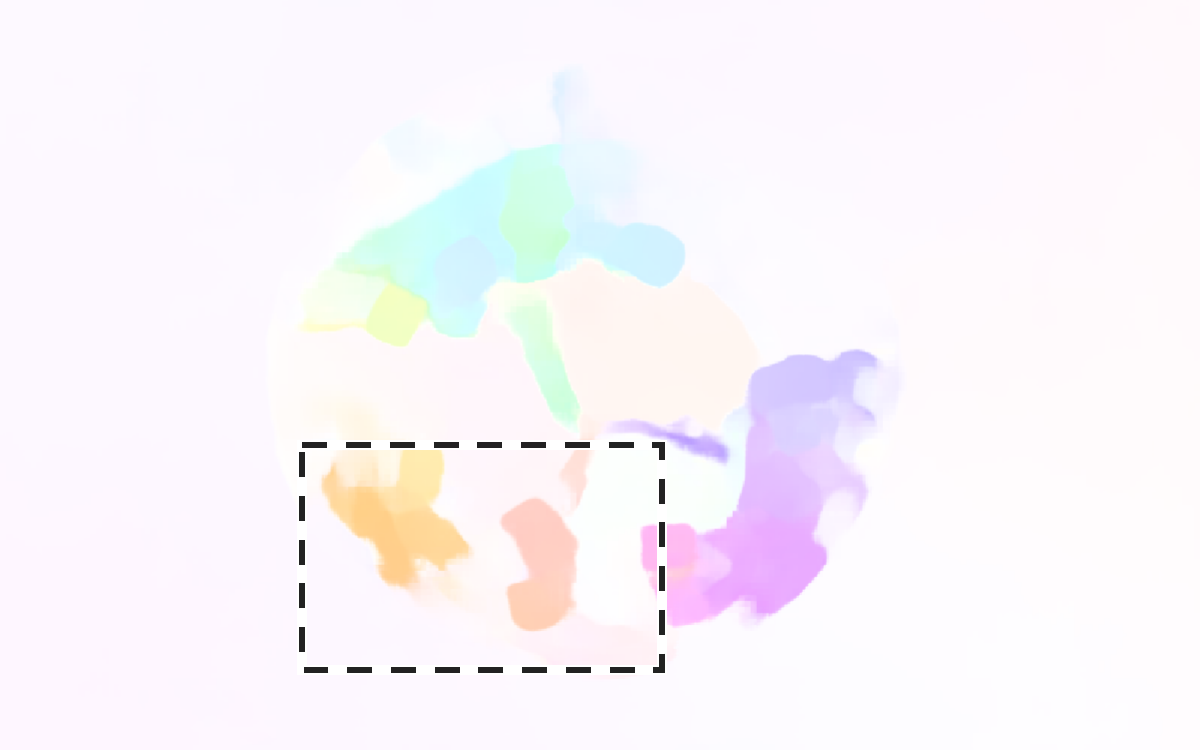} &
    \includegraphics[width=0.1041\linewidth]{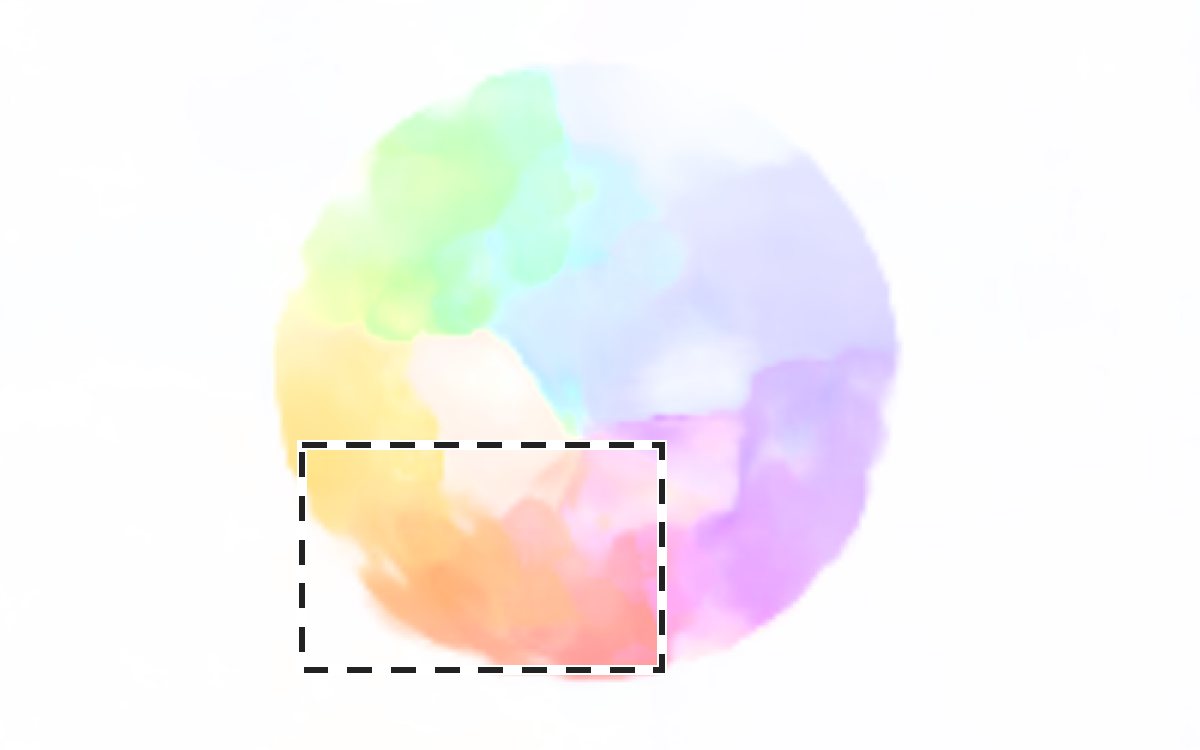} &
    \includegraphics[width=0.1041\linewidth]{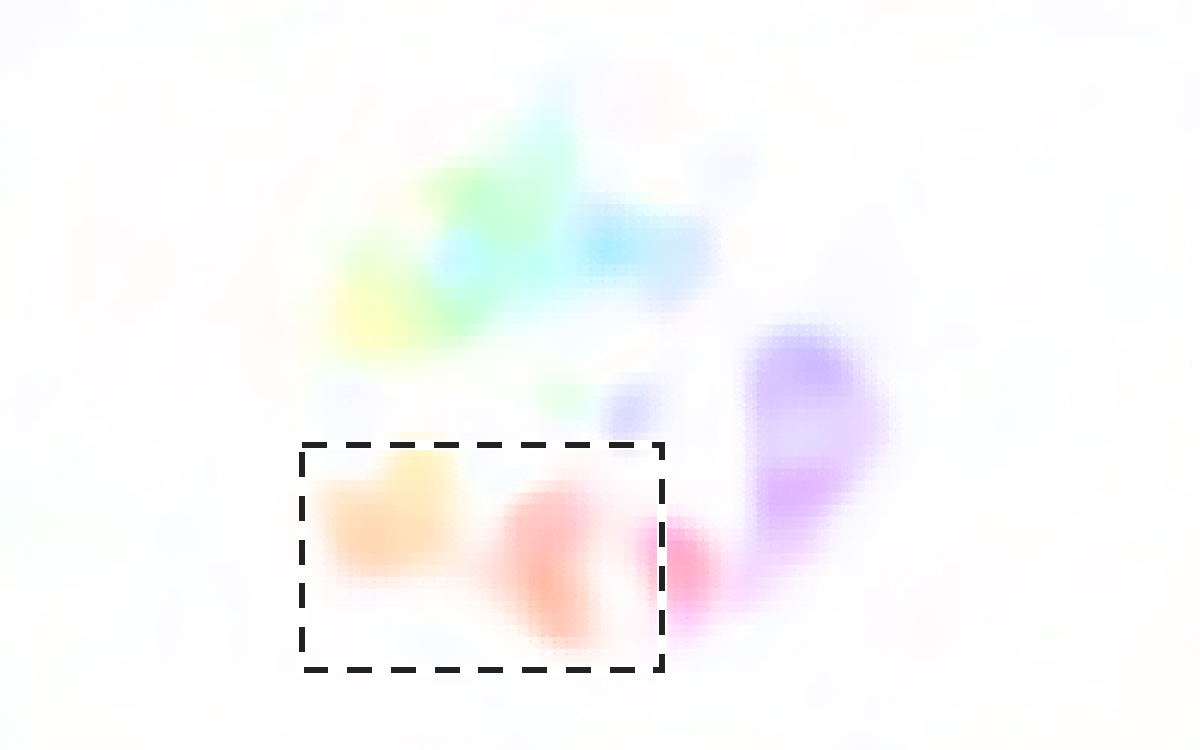} &
    \includegraphics[width=0.1041\linewidth]{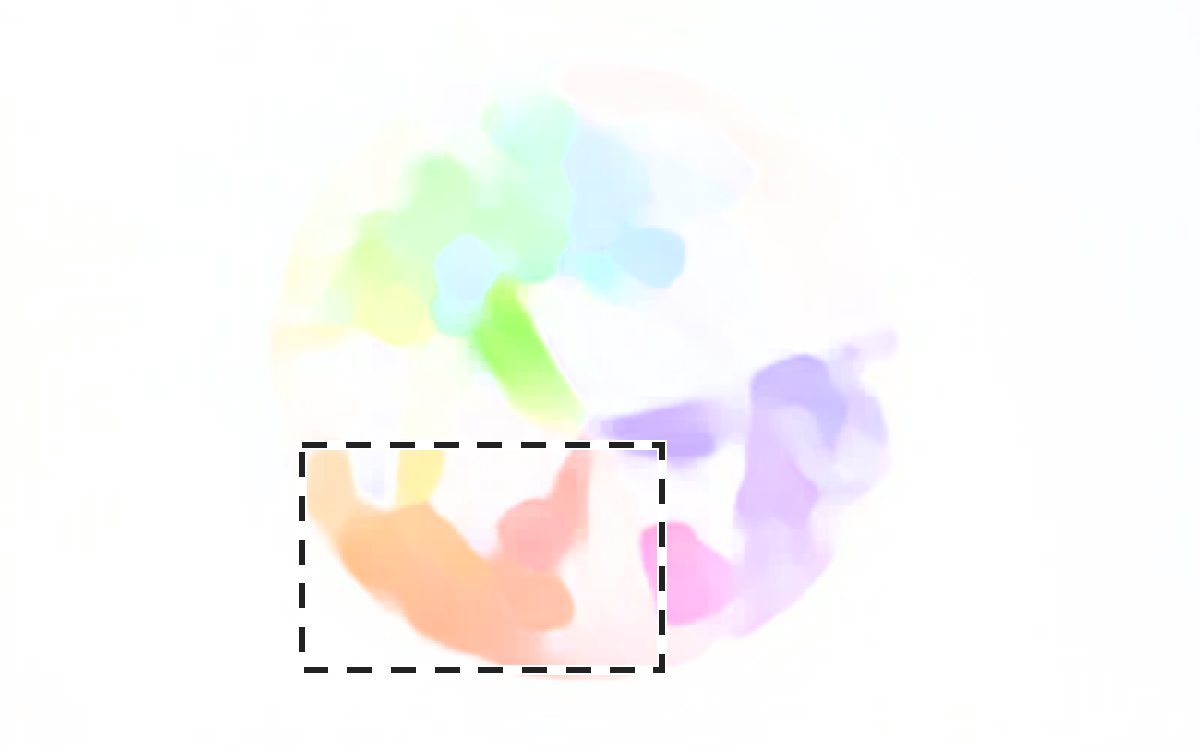} &
    \includegraphics[width=0.1041\linewidth]{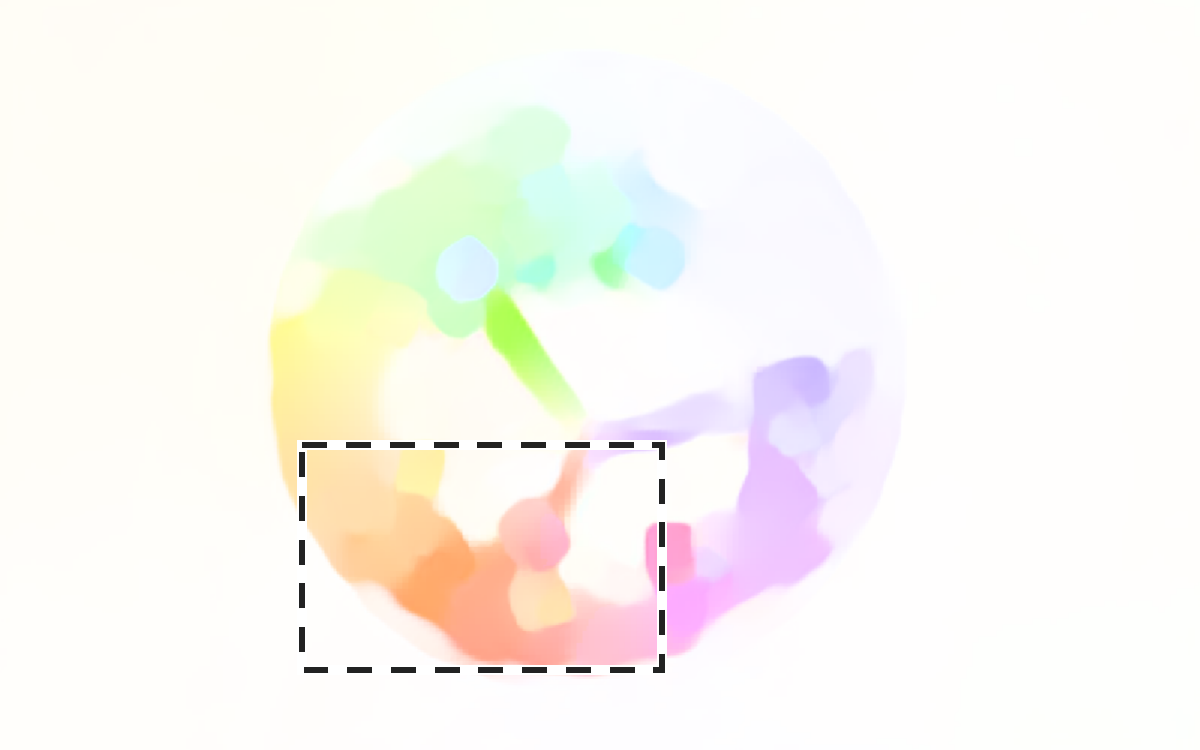} &
    \includegraphics[width=0.1041\linewidth]{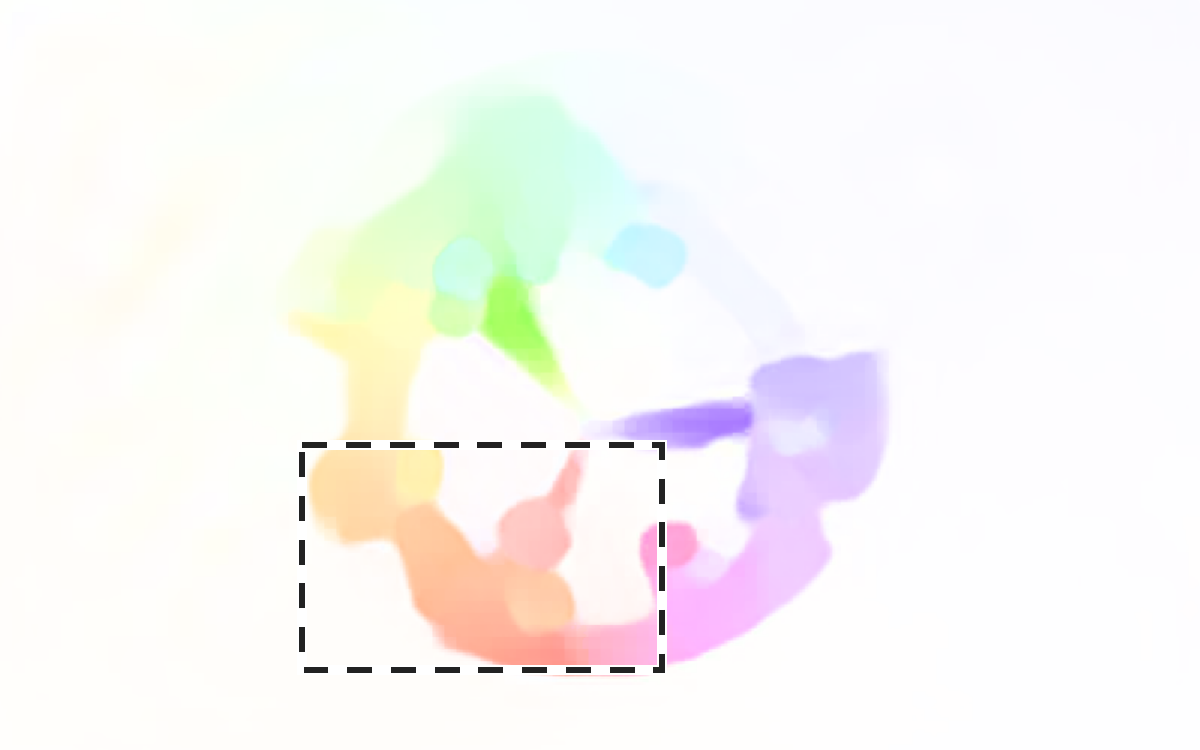} &
    \includegraphics[width=0.1041\linewidth]{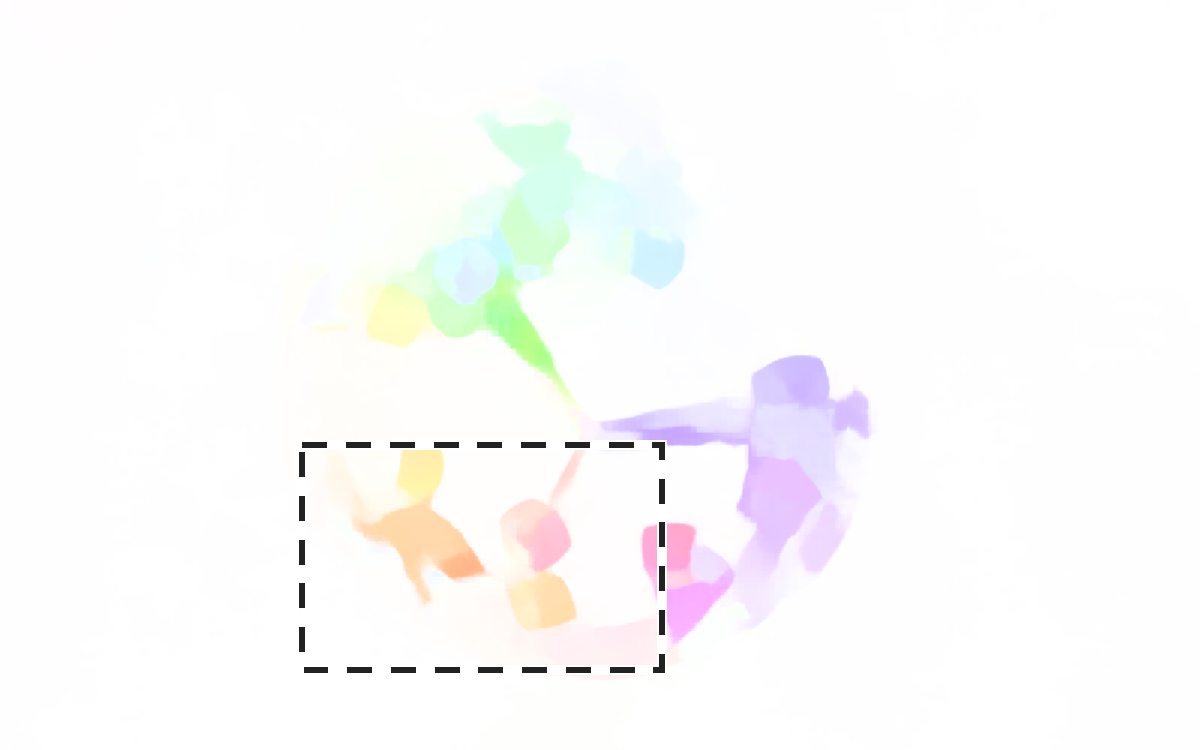} &
    \includegraphics[width=0.1041\linewidth]{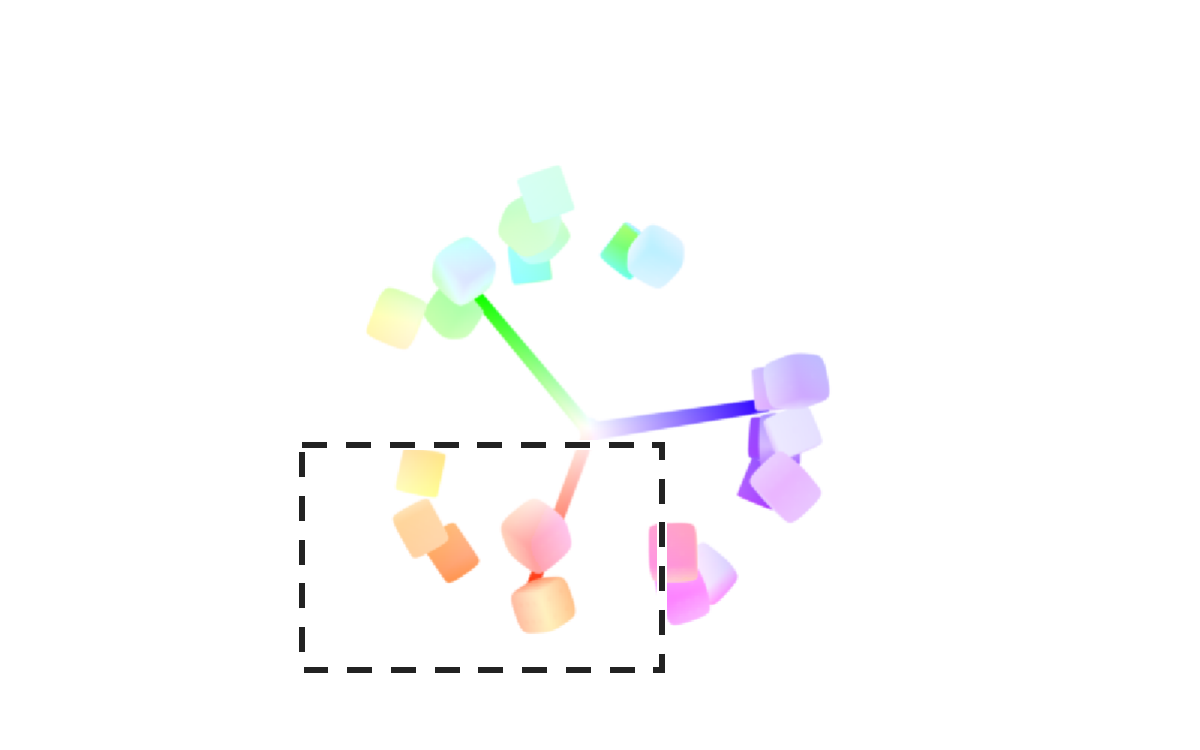} \\
    
    \scriptsize Dice ($\Delta t=10$) & \scriptsize 0.750 / 32.57 & \scriptsize 1.164 / 32.54 & \scriptsize 0.536 / 19.36 & \scriptsize 0.634 / 21.12 & \scriptsize 0.907 / 37.65 & \scriptsize 0.821 / 23.50 & \scriptsize \textbf{0.320 / 11.83} & \scriptsize Ground truth \\
    
    \includegraphics[width=0.1041\linewidth]{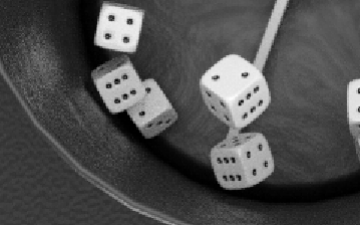} &
    \includegraphics[width=0.1041\linewidth]{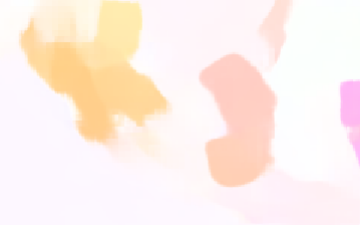} &
    \includegraphics[width=0.1041\linewidth]{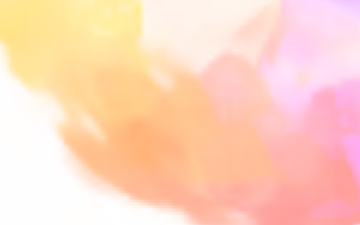} &
    \includegraphics[width=0.1041\linewidth]{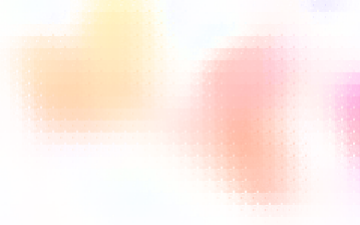} &
    \includegraphics[width=0.1041\linewidth]{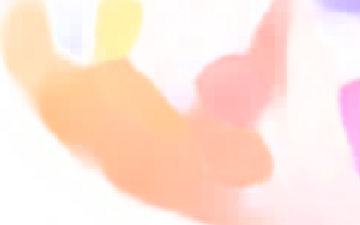} &
    \includegraphics[width=0.1041\linewidth]{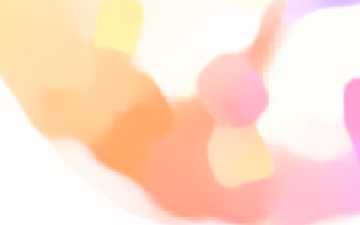} &
    \includegraphics[width=0.1041\linewidth]{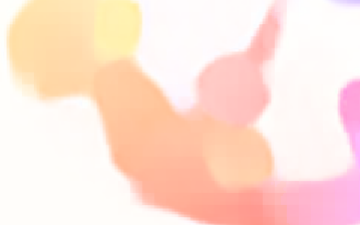} &
    \includegraphics[width=0.1041\linewidth]{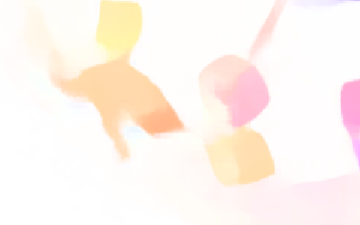} &
    \includegraphics[width=0.1041\linewidth]{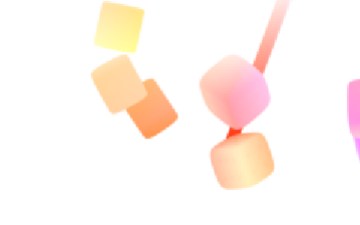} \\
    
    \includegraphics[width=0.1041\linewidth]{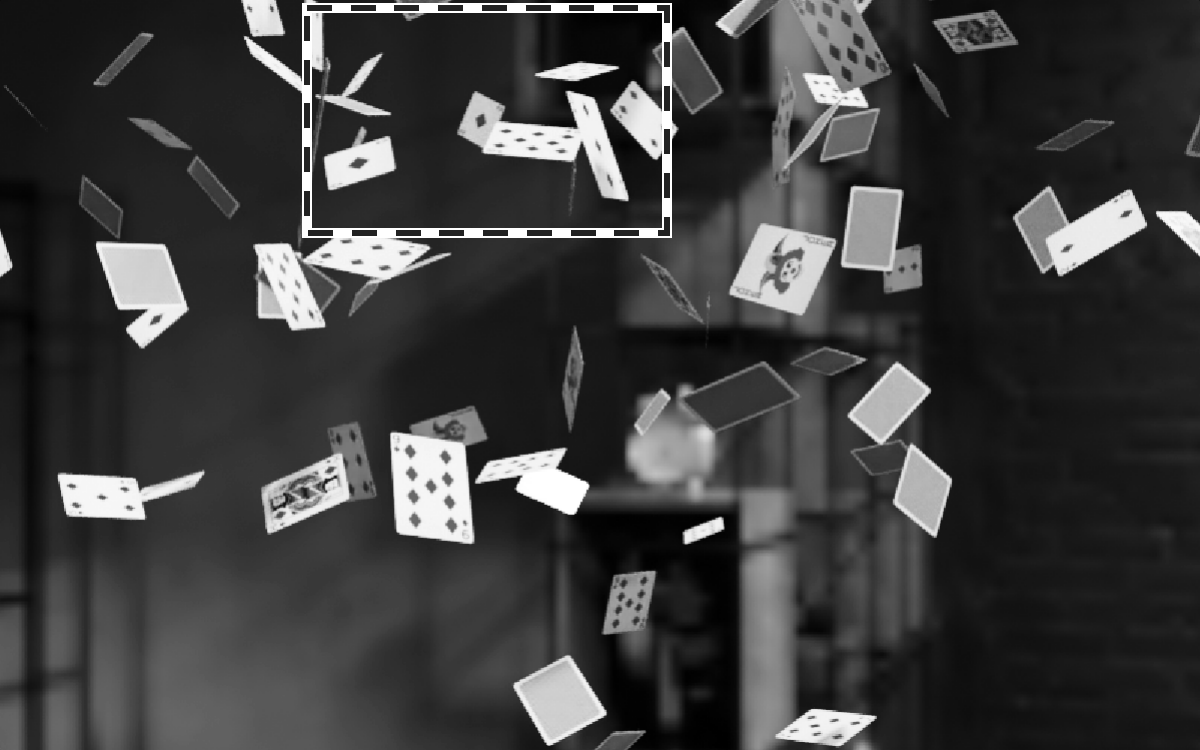} &
    \includegraphics[width=0.1041\linewidth]{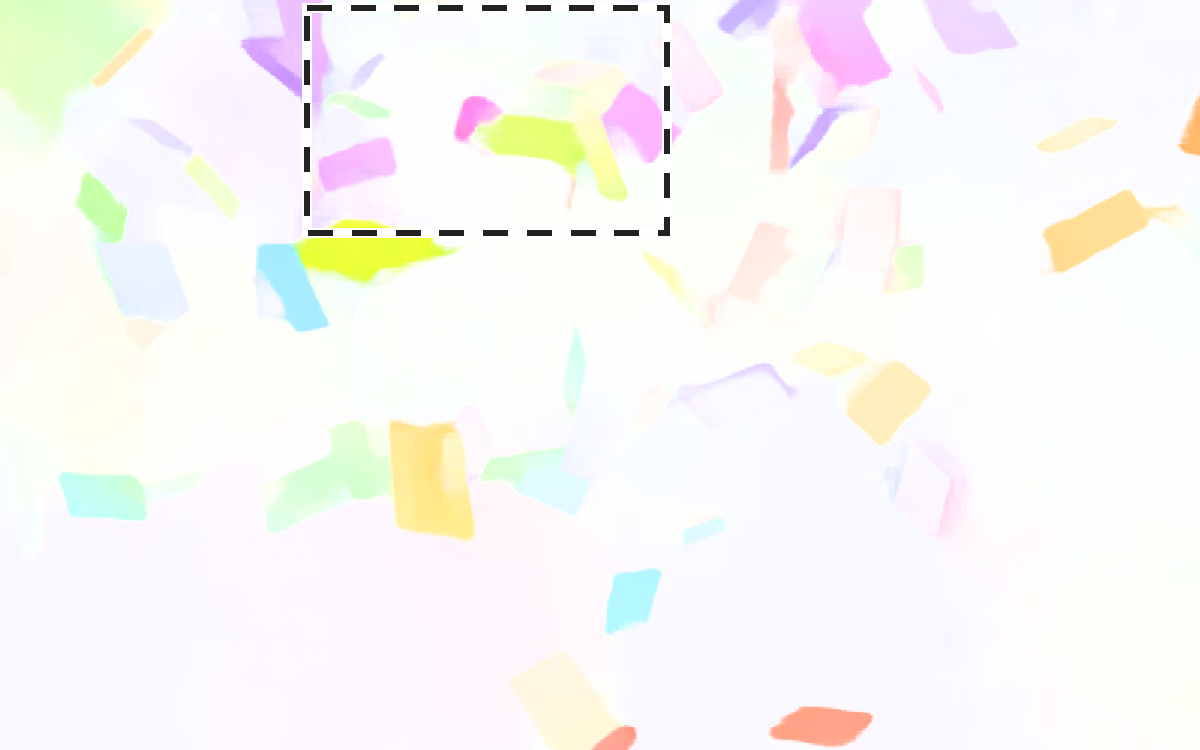} &
    \includegraphics[width=0.1041\linewidth]{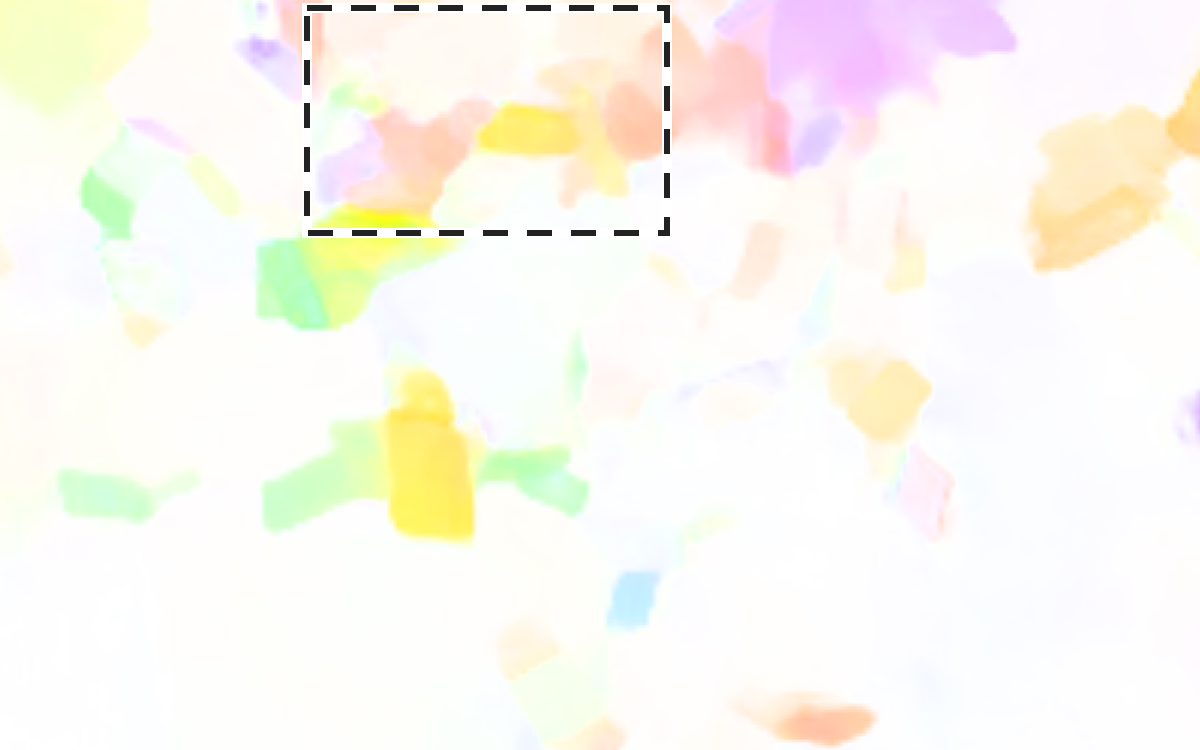} &
    \includegraphics[width=0.1041\linewidth]{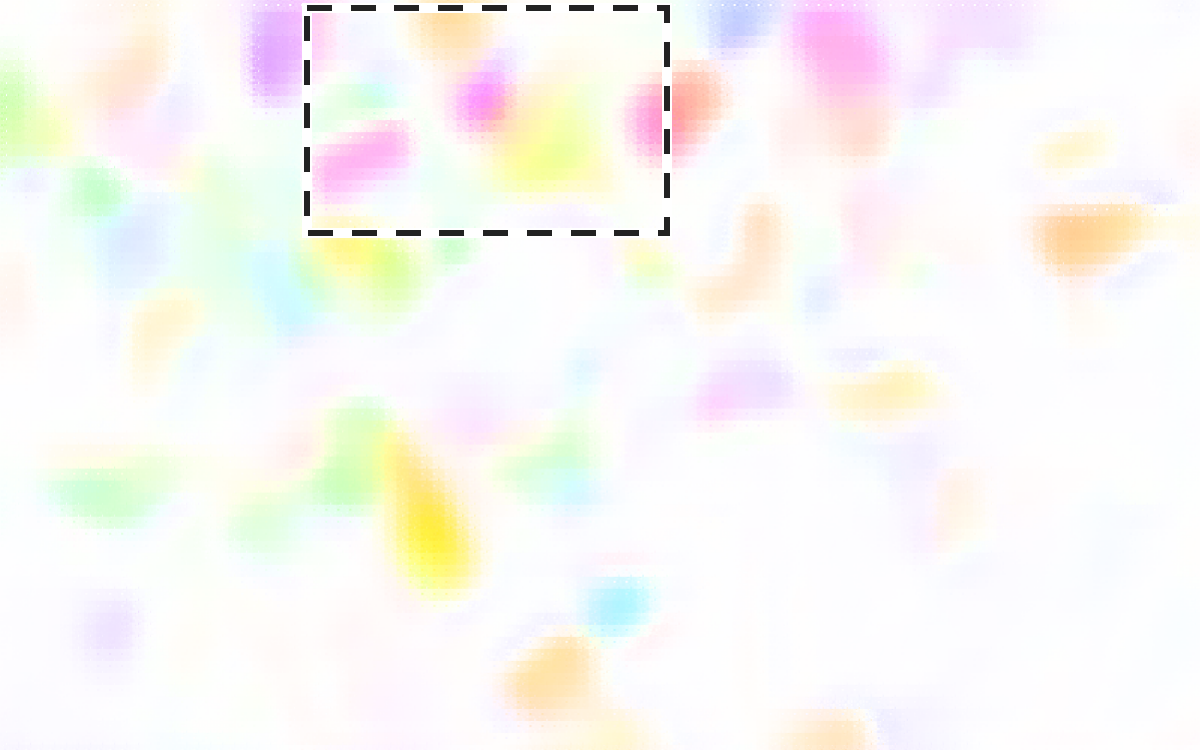} &
    \includegraphics[width=0.1041\linewidth]{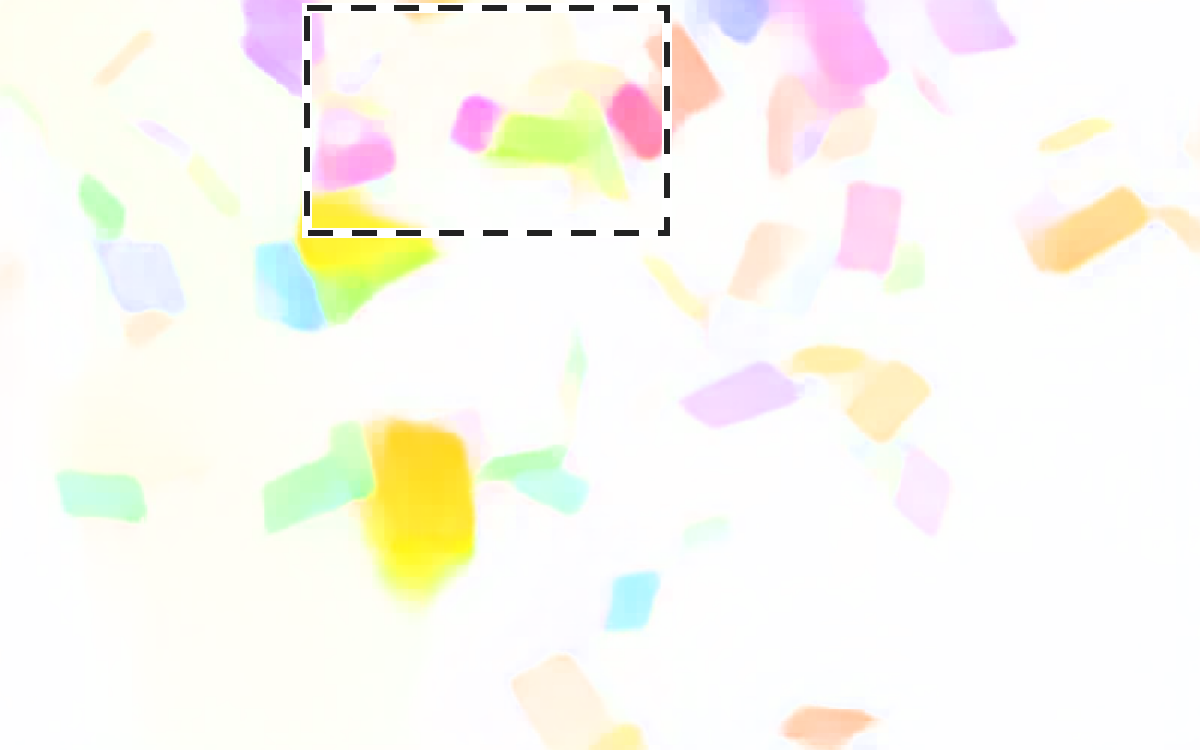} &
    \includegraphics[width=0.1041\linewidth]{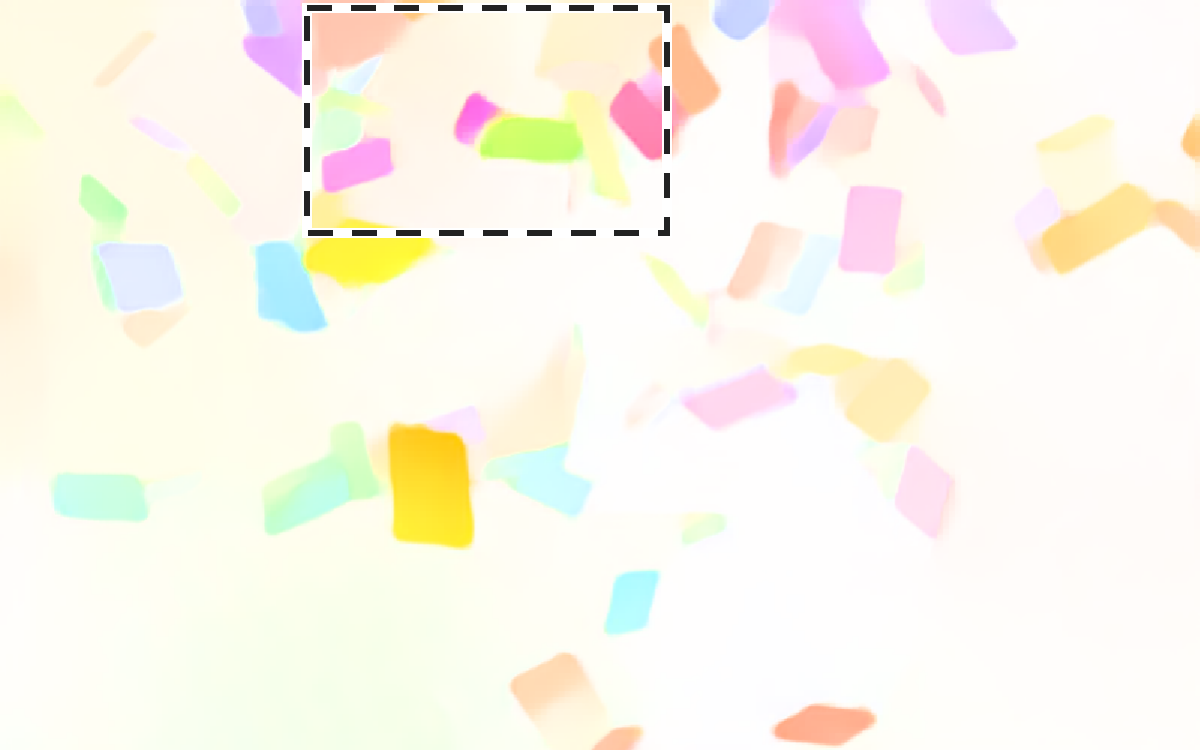} &
    \includegraphics[width=0.1041\linewidth]{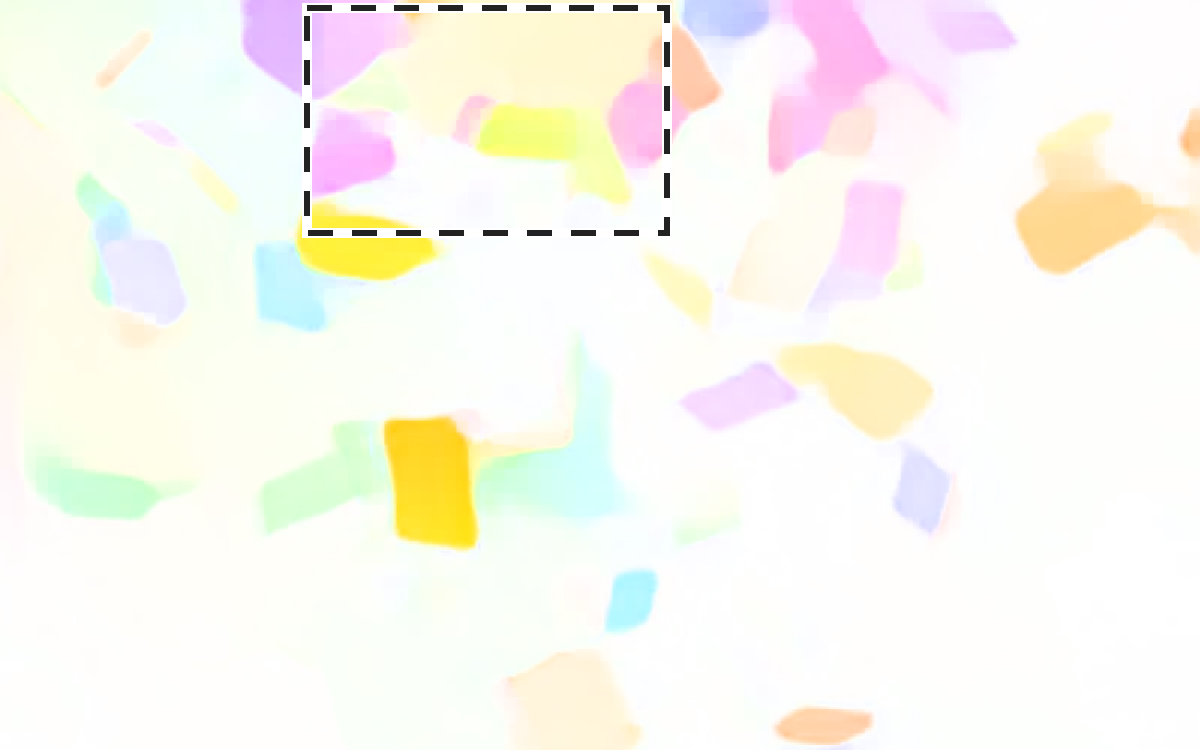} &
    \includegraphics[width=0.1041\linewidth]{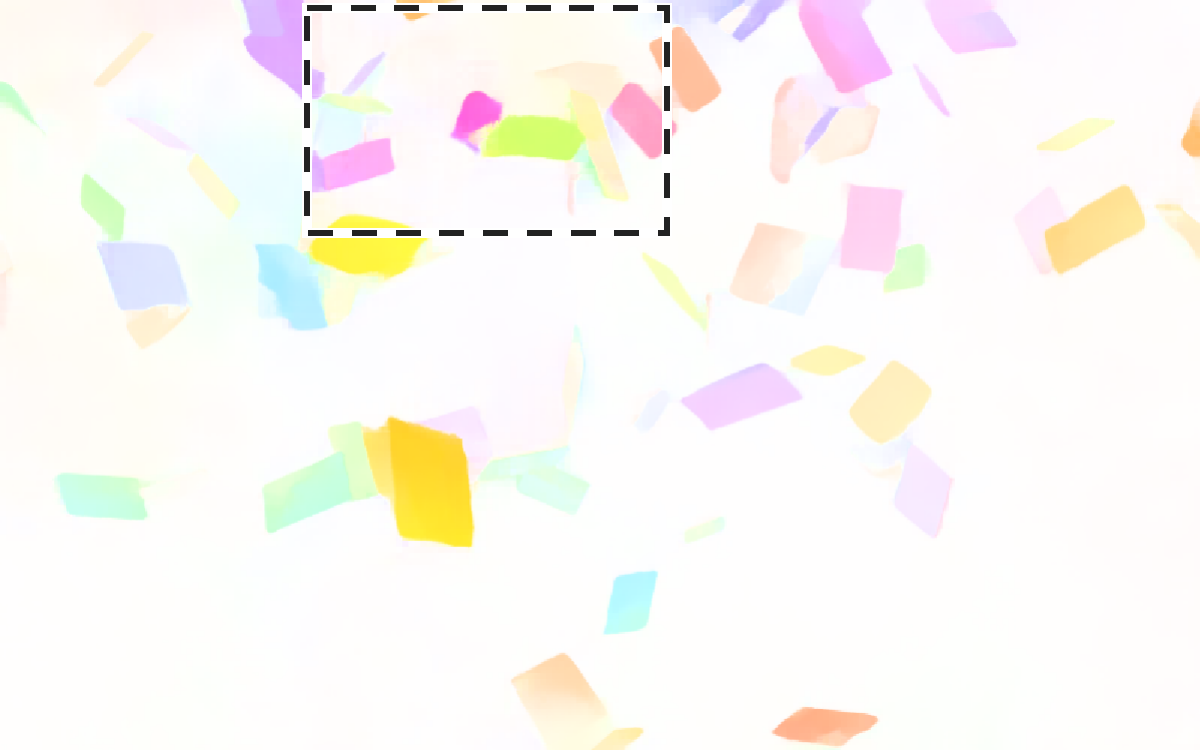} &
    \includegraphics[width=0.1041\linewidth]{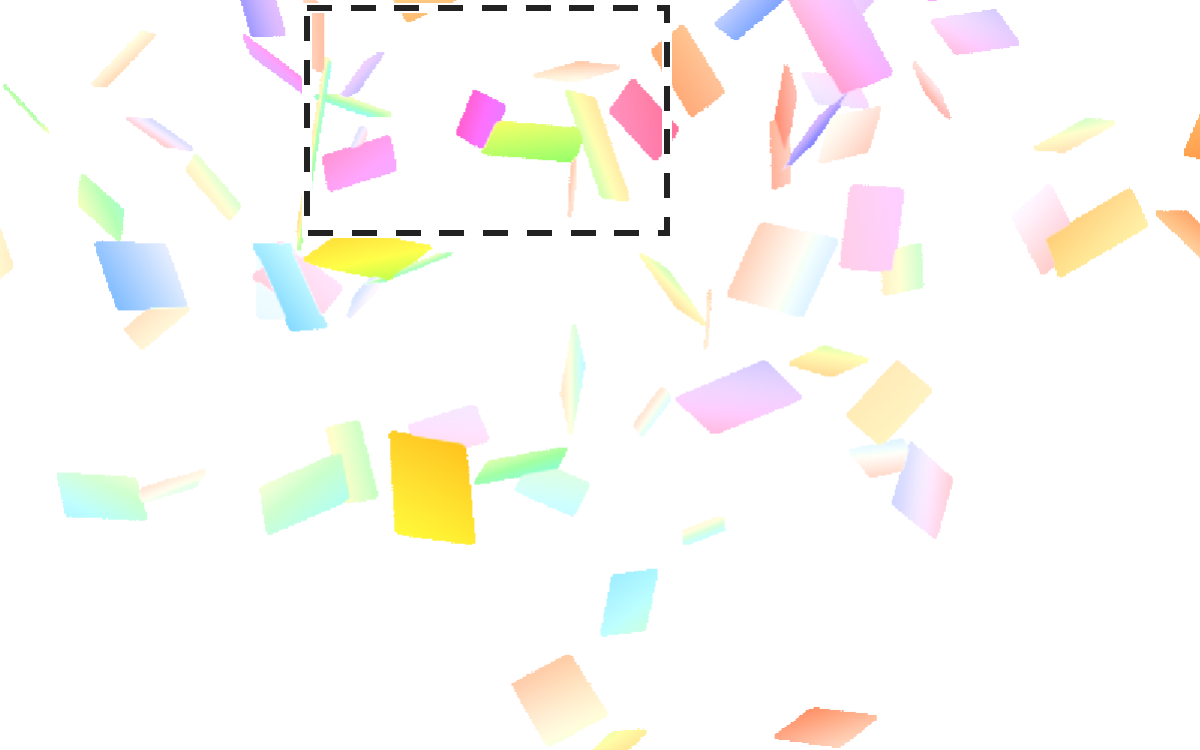} \\
    
    \scriptsize Poker ($\Delta t=10$) & \scriptsize 0.619 / 20.86 & \scriptsize 0.776 / 31.21 & \scriptsize 0.744 / 35.56 & \scriptsize 0.539 / 24.35 & \scriptsize 0.704 / 47.05 & \scriptsize 0.700 / 38.46 & \scriptsize \textbf{0.393 / 16.46} & \scriptsize Ground truth \\
    
    \includegraphics[width=0.1041\linewidth]{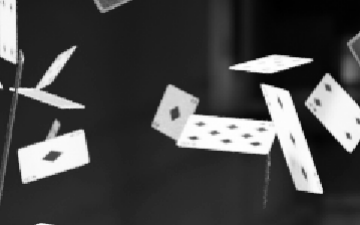} &
    \includegraphics[width=0.1041\linewidth]{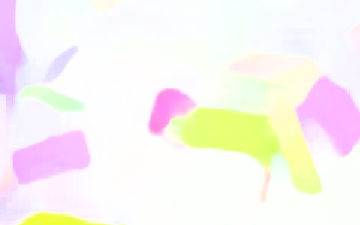} &
    \includegraphics[width=0.1041\linewidth]{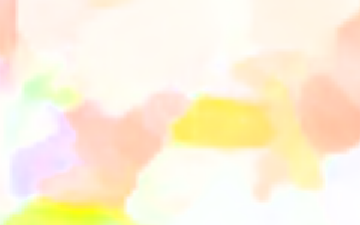} &
    \includegraphics[width=0.1041\linewidth]{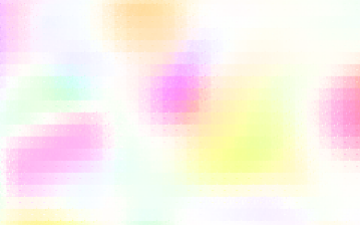} &
    \includegraphics[width=0.1041\linewidth]{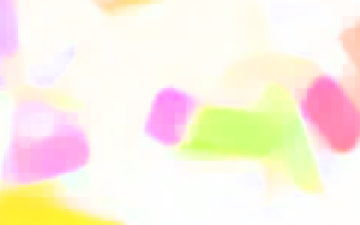} &
    \includegraphics[width=0.1041\linewidth]{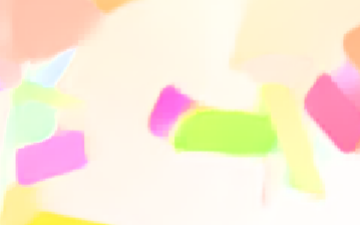} &
    \includegraphics[width=0.1041\linewidth]{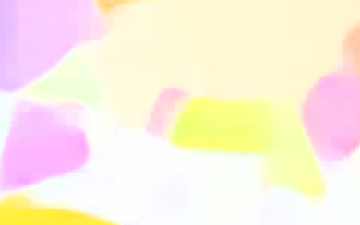} &
    \includegraphics[width=0.1041\linewidth]{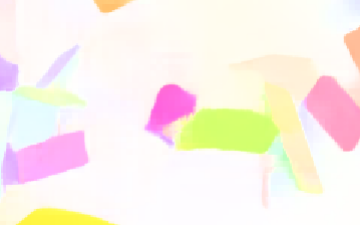} &
    \includegraphics[width=0.1041\linewidth]{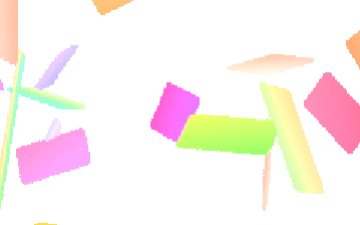} \\
    
    \end{tabular}
    \caption{\textbf{Qualitative comparison on PHM.}
    Numbers report whole-image AEPE (pixels) / PO (\%).
    Flow maps within each example share a color scale.
    Dashed boxes mark the regions enlarged below.}
    \label{fig:phm_qualitative_all_methods_zoom}
\end{figure*}

\begin{figure*}[!t]
    \centering
    \setlength{\tabcolsep}{1pt}
    \begin{tabular}{@{}cccccccc@{}}
    \scriptsize Scene & \scriptsize EMatch & \scriptsize EEMFlow+ & \scriptsize Spike2Flow & \scriptsize Spike2Flow++ & \scriptsize SIV & \scriptsize HiST-SFlow & \scriptsize \textbf{TIDE-Flow (Ours)} \\
    \includegraphics[width=0.1179\linewidth]{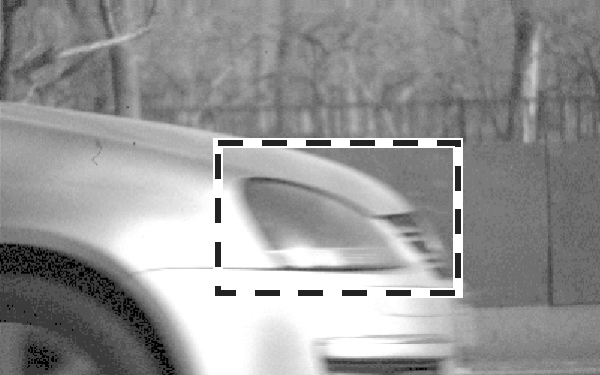} & \includegraphics[width=0.1179\linewidth]{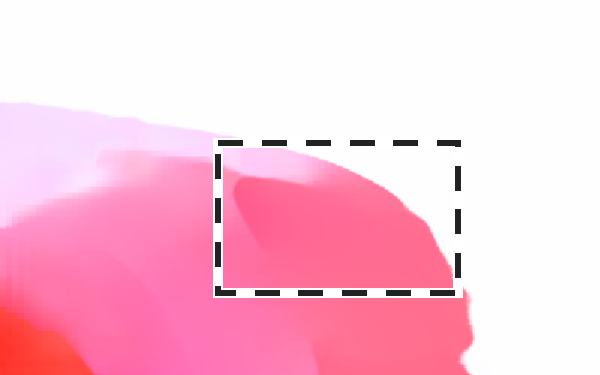} & \includegraphics[width=0.1179\linewidth]{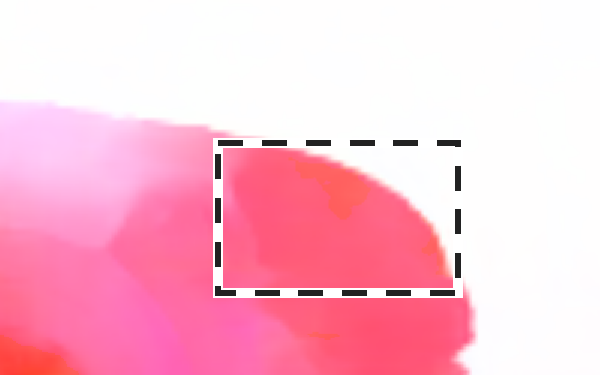} & \includegraphics[width=0.1179\linewidth]{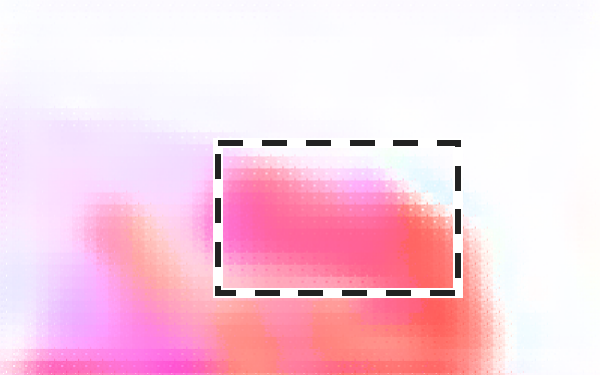} & \includegraphics[width=0.1179\linewidth]{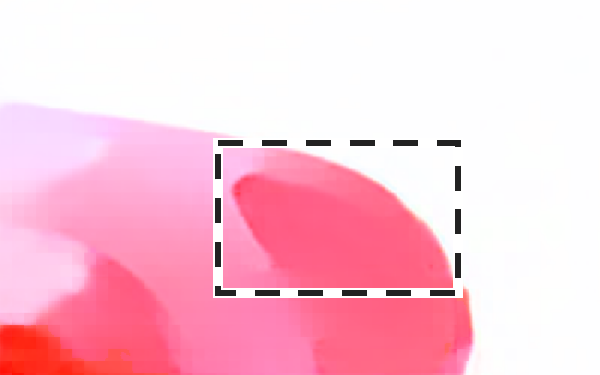} & \includegraphics[width=0.1179\linewidth]{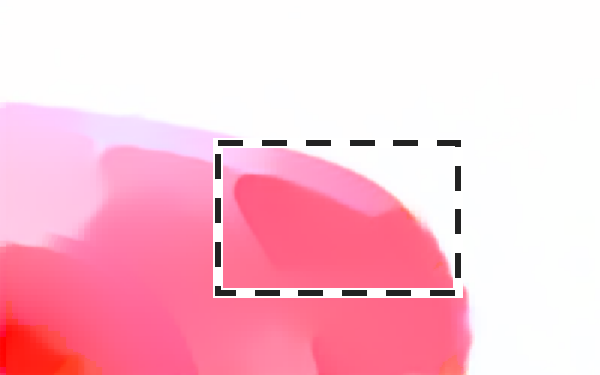} & \includegraphics[width=0.1179\linewidth]{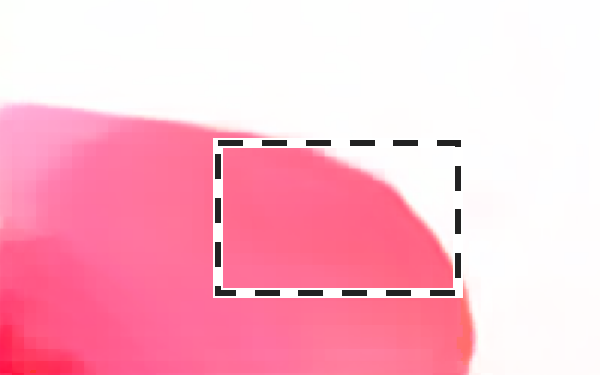} & \includegraphics[width=0.1179\linewidth]{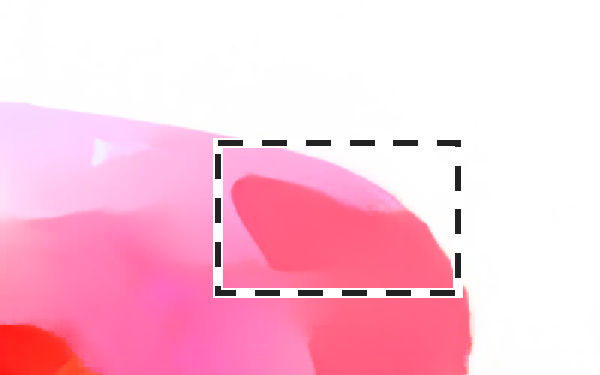} \\
    \includegraphics[width=0.1179\linewidth]{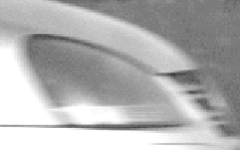} & \includegraphics[width=0.1179\linewidth]{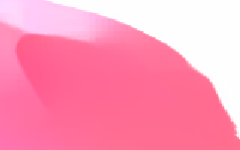} & \includegraphics[width=0.1179\linewidth]{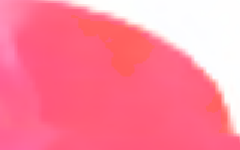} & \includegraphics[width=0.1179\linewidth]{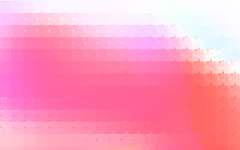} & \includegraphics[width=0.1179\linewidth]{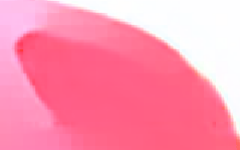} & \includegraphics[width=0.1179\linewidth]{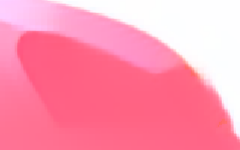} & \includegraphics[width=0.1179\linewidth]{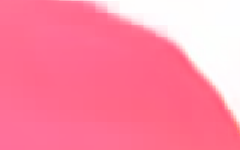} & \includegraphics[width=0.1179\linewidth]{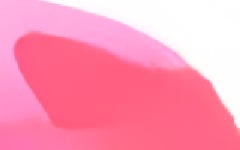} \\
    \includegraphics[width=0.1179\linewidth]{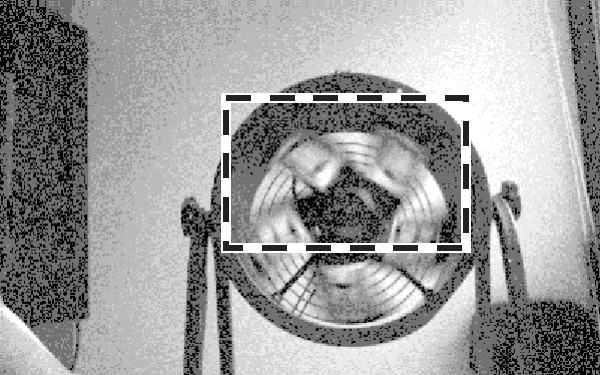} & \includegraphics[width=0.1179\linewidth]{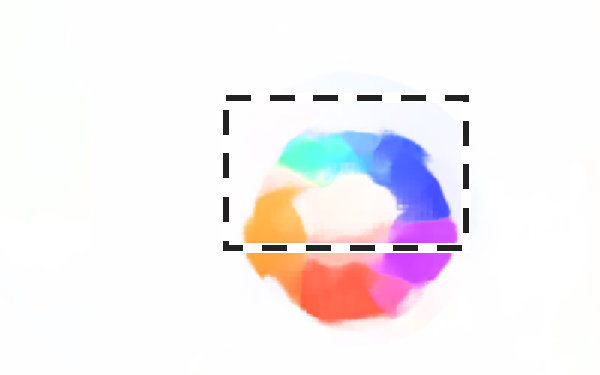} & \includegraphics[width=0.1179\linewidth]{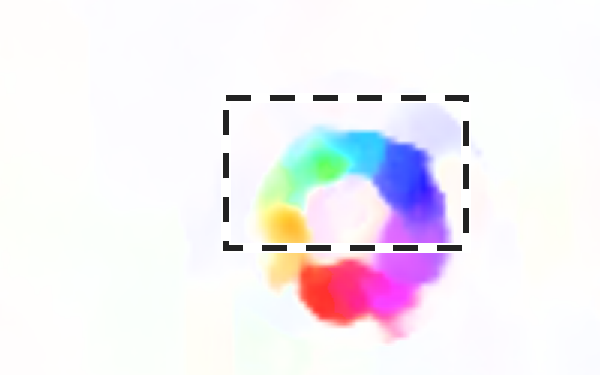} & \includegraphics[width=0.1179\linewidth]{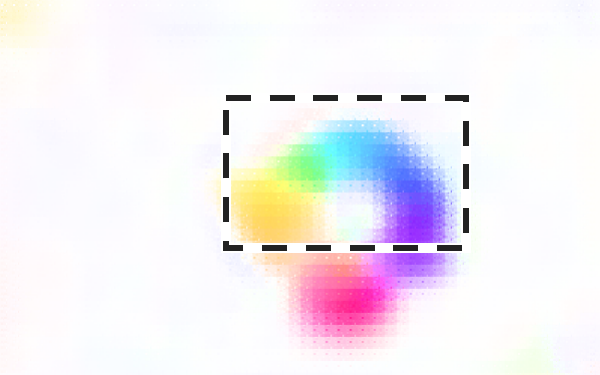} & \includegraphics[width=0.1179\linewidth]{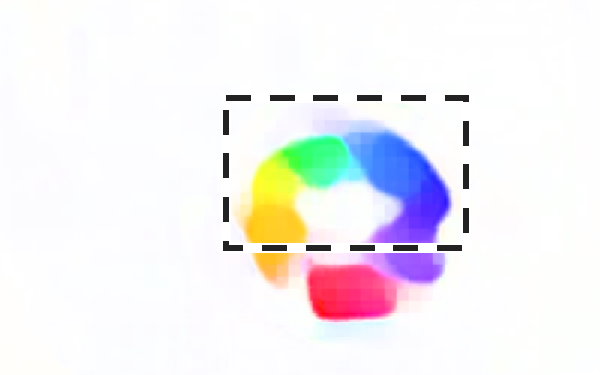} & \includegraphics[width=0.1179\linewidth]{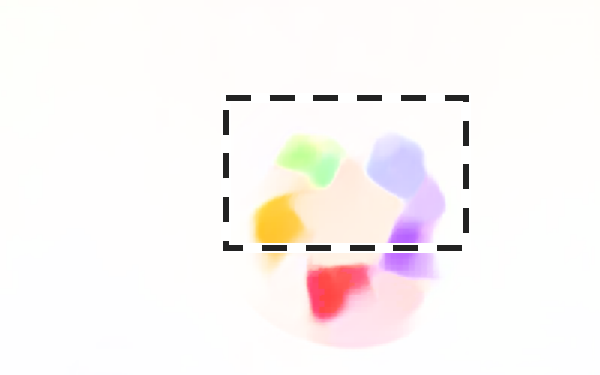} & \includegraphics[width=0.1179\linewidth]{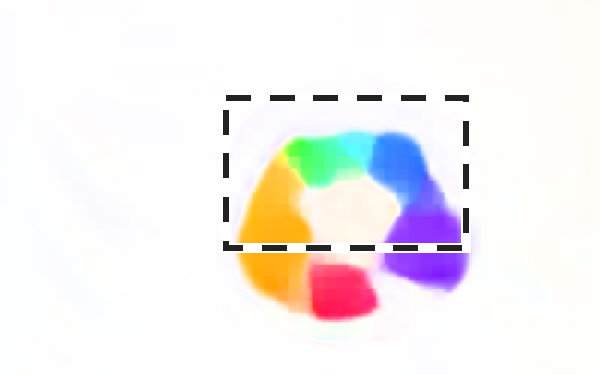} & \includegraphics[width=0.1179\linewidth]{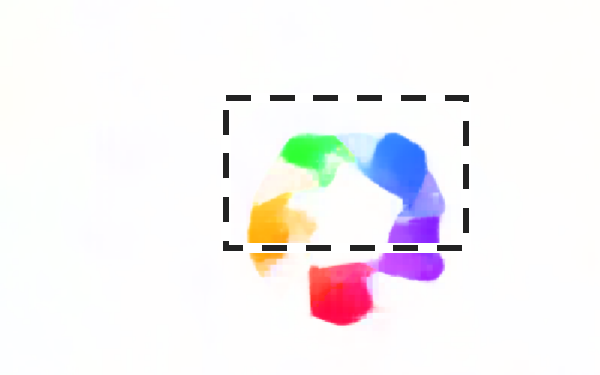} \\
    \includegraphics[width=0.1179\linewidth]{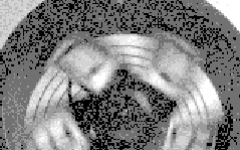} & \includegraphics[width=0.1179\linewidth]{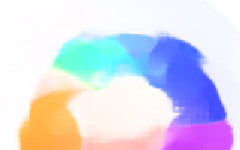} & \includegraphics[width=0.1179\linewidth]{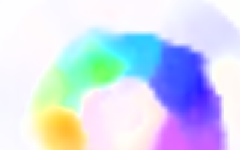} & \includegraphics[width=0.1179\linewidth]{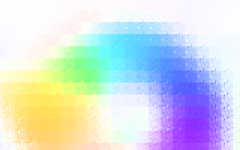} & \includegraphics[width=0.1179\linewidth]{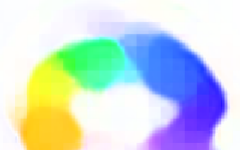} & \includegraphics[width=0.1179\linewidth]{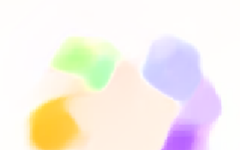} & \includegraphics[width=0.1179\linewidth]{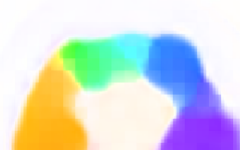} & \includegraphics[width=0.1179\linewidth]{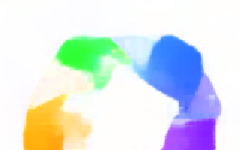} \\
    \end{tabular}
\caption{Qualitative comparison on real-captured spike sequences at $\Delta t=20$ Spike frames. The first column is a 25-frame temporal-mean visualization of the source Spike window. Dashed boxes indicate the same regions enlarged below.}
\label{fig:figure/real_qualitative_all_methods_zoom}
\end{figure*}

\vspace{-10pt}
\paragraph{Qualitative Comparisons.}
\cref{fig:phm_qualitative_all_methods_zoom} presents qualitative comparisons on the PHM benchmark. Compared with existing methods, TIDE-Flow produces more accurate motion boundaries and better preserves moving object structures. In the Dice and Poker scenes, competing methods often exhibit blurred motion regions, distorted object shapes, or inaccurate flow transitions, whereas our predictions show clearer object separation and more consistent motion patterns that closely resemble the ground truth. We further evaluate the generalization ability of TIDE-Flow on real-captured spike sequences in \cref{fig:figure/real_qualitative_all_methods_zoom}. Without fine-tuning, our method produces coherent flow fields with clearer motion boundaries around the vehicle and rotating object, while exhibiting fewer artifacts in the enlarged regions. These results demonstrate the effectiveness of TIDE-Flow in capturing fine-grained motion information from spike streams.

\subsection{Ablation Studies}

Unless otherwise specified, we train the ablation variants on SPIFT for 50K iterations and evaluate them on PHM.

\begin{table}[!t]
    \centering
    \footnotesize
    \caption{\textbf{Component ablation on PHM.}
    Norm.\ denotes matching-feature normalization before correlation.
    Boldface marks the best value in each metric column.}
    \begin{tabularx}{\linewidth}{@{}c|ccc|CC|CC@{}}
        \toprule[1pt]
        \multirow{2}{*}{Index} & \multicolumn{3}{c|}{Settings} & \multicolumn{2}{c|}{$\Delta t=10$} & \multicolumn{2}{c}{$\Delta t=20$} \\
        \cmidrule(lr){2-4} \cmidrule(lr){5-6} \cmidrule(lr){7-8}
        & TIE & Norm. & DEE & AEPE & PO\% & AEPE & PO\% \\
        \midrule
        (A) & \xmark & \xmark & \xmark & 0.7066 & 16.15 & 1.4418 & 21.00 \\
        (B) & \cmark & \xmark & \xmark & 0.6620 & 14.94 & 1.3642 & 19.28 \\
        (C) & \cmark & \cmark & \xmark & 0.6793 & 16.11 & 1.3303 & 19.73 \\
        (D) & \cmark & \xmark & \cmark & 0.6496 & 15.13 & 1.3057 & 19.07 \\
        \rowcolor{gray!15}
        (E) & \cmark & \cmark & \cmark & \textbf{0.6401} & \textbf{14.44} & \textbf{1.2786} & \textbf{17.83} \\
        \bottomrule[1pt]
    \end{tabularx}
    \label{tab:ablation_components}
\end{table}

\vspace{-10pt}
\paragraph{Ablations for Modules.}
We conduct ablation studies to evaluate the effectiveness of the proposed components, as shown in~\cref{tab:ablation_components}. Comparing (A) and (B), introducing TIE reduces AEPE by 6.31\% and 5.38\% for $\Delta t=10$ and $\Delta t=20$, respectively, demonstrating the benefit of preserving temporal integration evidence. Further incorporating DEE in (D) improves AEPE under both intervals, indicating the effectiveness of encoding displacement-aware matching information. Although correlation normalization alone yields mixed improvements in (C), combining it with DEE in (E) achieves the best performance across all metrics, with AEPE reduced to 0.6401 and 1.2786 and PO\% to 14.44 and 17.83, respectively. These results demonstrate the effectiveness of the proposed components and their complementary contributions to optical flow estimation.

\begin{table}[!t]
    \centering
    \footnotesize
    \caption{\textbf{Temporal input and encoder ablation on PHM.} Fine and Rate denote fine-grained spike evidence and temporally averaged firing rates, respectively. Boldface marks the best value in each metric column.}
    \begin{tabularx}{\linewidth}{@{}c l|CC|CC@{}}
        \toprule[1pt]
        \multicolumn{2}{c|}{Settings} & \multicolumn{2}{c|}{$\Delta t=10$} & \multicolumn{2}{c}{$\Delta t=20$} \\
        \cmidrule(lr){1-2} \cmidrule(lr){3-4} \cmidrule(lr){5-6}
        Evidence & Encoder & AEPE & PO\% & AEPE & PO\% \\
        \midrule
        Rate & Original & 0.7591 & 21.26 & 1.4991 & 27.46 \\
        Rate & TIE Stem & 0.7843 & 22.60 & 1.5630 & 31.99 \\
        Fine & Original & 0.7084 & 17.85 & 1.4399 & 24.30 \\
        \rowcolor{gray!15}
        Fine & TIE Stem & \textbf{0.6681} & \textbf{15.47} & \textbf{1.3343} & \textbf{19.13} \\
        \bottomrule[1pt]
    \end{tabularx}
    \label{tab:ablation_tie}
\end{table}

\vspace{-10pt}
\paragraph{Temporal Evidence Encoding.}
We compare the TIE stem and the original stem with fine-grained spike evidence (Fine) and temporally averaged firing rates (Rate), holding the matching configuration fixed and omitting explicit interval cues. In \cref{tab:ablation_tie}, the TIE stem lowers AEPE by 5.69\% and 7.33\% with Fine at the two intervals, while increasing both errors with Rate. The reversal links the benefit of the stem to the information available in its input.

\begin{table}[!t]
    \footnotesize
    \centering
    \caption{\textbf{Temporal evidence ablation on PHM.} Boldface marks the best value in each metric column.}
    \begin{tabularx}{\linewidth}{@{}l|CC|CC@{}}
        \toprule[1pt]
        \multirow{2}[2]{*}{Variant} & \multicolumn{2}{c|}{$\Delta t=10$} & \multicolumn{2}{c}{$\Delta t=20$} \\
        \cmidrule(lr){2-3} \cmidrule(lr){4-5} & AEPE  $\downarrow$ & PO\%  $\downarrow$ & AEPE  $\downarrow$ & PO\%  $\downarrow$ \\
        \midrule
        \textit{w/o} Raw Spike & 0.7386 & 18.87 & 1.6830 & 26.19 \\
        \textit{w/o} Rate & 0.6829 & 15.20 & 1.3359 & 18.67 \\
        \textit{w/o} Moment & 0.6650 & 16.49 & 1.3189 & 20.30 \\
        \textit{w/o} Interval & 0.6843 & 15.86 & 1.3374 & 18.91 \\
        \rowcolor{gray!15}
        Full TIE & \textbf{0.6401} & \textbf{14.44} & \textbf{1.2786} & \textbf{17.83} \\
        \bottomrule[1pt]
    \end{tabularx}
    \label{tab:tie_temporal_components}
\end{table}

\vspace{-10pt}
\paragraph{Temporal Evidence Components.}
Removing raw spikes produces the largest increase in AEPE, from 0.6401 to 0.7386 pixels at $\Delta t=10$ and from 1.2786 to 1.6830 pixels at $\Delta t=20$ (\cref{tab:tie_temporal_components}). Removing the interval channels also increases AEPE at both intervals, by 0.0442 and 0.0588 pixels, respectively, with the raw spike sequence still available. All four removals increase both metrics, showing how the temporal summaries complement the observed firing sequence.

\begin{table}[!t]
    \centering
    \footnotesize
    \caption{\textbf{Displacement evidence ablation on PHM.} Boldface marks the best value in each metric column.}
    \begin{tabularx}{\linewidth}{@{}l|CC|CC@{}}
        \toprule[1pt]
        \multirow{2}[2]{*}{Variant} &
        \multicolumn{2}{c|}{$\Delta t=10$} &
        \multicolumn{2}{c}{$\Delta t=20$} \\
        \cmidrule(lr){2-3} \cmidrule(lr){4-5}
        & AEPE & PO\% & AEPE & PO\% \\
        \midrule
        CostMLP
            & 0.6436 & 14.50 & 1.3429 & 18.21 \\
        \textit{w/o} Statistics
            & 0.6674 & 15.61 & 1.3615 & 19.42 \\
        MLP + Statistics
            & 0.6670 & 15.49 & 1.3223 & 19.35 \\
        \rowcolor{gray!15}
        Full DEE
            & \textbf{0.6401} & \textbf{14.44}
            & \textbf{1.2786} & \textbf{17.83} \\
        \bottomrule[1pt]
    \end{tabularx}
    \label{tab:ablation_dee}
\end{table}

\vspace{-10pt}
\paragraph{Displacement Evidence Encoding.}
We evaluate the effectiveness of DEE by comparing it with a capacity-comparable CostMLP and two variants that remove the explicit statistics or replace the structured grid encoder with an MLP. As shown in \cref{tab:ablation_dee}, the full DEE achieves the best performance across all metrics. Removing the statistics or replacing the grid encoder consistently degrades accuracy, confirming the complementary benefits of explicit displacement statistics and structured candidate-grid encoding. The improvements are particularly evident at $\Delta t=20$, demonstrating the effectiveness of DEE under larger temporal displacements.

\begin{figure}[!t]
    \centering
    \includegraphics[width=\linewidth]{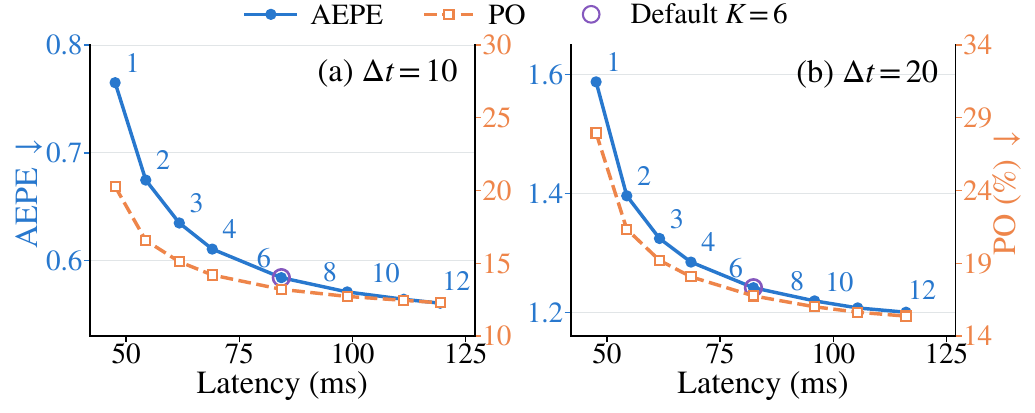}
    \caption{\textbf{Effect of recurrent updates on PHM.}
    AEPE, PO, and inference latency are reported at both temporal intervals.}
    \label{fig:iteration}
\end{figure}

\vspace{-10pt}
\paragraph{Number of Recurrent Updates.}
We vary the number of inference updates for the model trained for 100K iterations. As shown in \cref{fig:iteration}, increasing the number of updates consistently reduces AEPE and PO\% under both temporal intervals, with substantial improvements during the early iterations. However, the accuracy gains gradually diminish beyond $K=6$, while inference latency continues to increase. We therefore adopt six recurrent updates to balance estimation accuracy and computational efficiency.

\begin{figure}[!t]
    \centering
    \includegraphics[width=\linewidth]{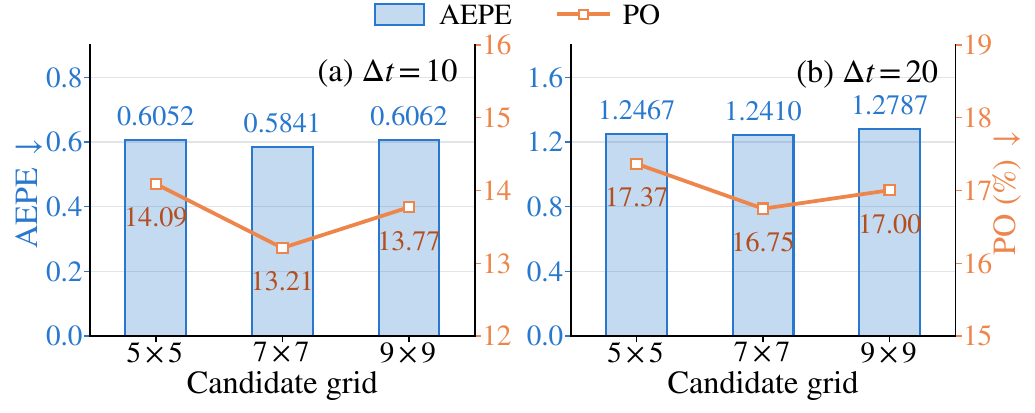}
    \caption{\textbf{Effect of candidate grid size on PHM.}
    AEPE and PO are compared for $5\times5$, $7\times7$, and $9\times9$ correlation neighborhoods.}
    \label{fig:grid}
\end{figure}

\vspace{-10pt}
\paragraph{Candidate Grid Size.}
We evaluate the effect of local candidate grid on DEE using $5\times5$, $7\times7$, and $9\times9$ correlation neighborhoods for 100K iterations. As shown in~\cref{fig:grid}, the $7\times7$ grid achieves the lowest AEPE and PO\% at $\Delta t=10$, while obtaining the lowest PO\% and comparable AEPE at $\Delta t=20$. Smaller neighborhoods may provide insufficient displacement context, whereas larger neighborhoods do not consistently improve matching accuracy. We therefore adopt a $7\times7$ candidate grid for its favorable overall performance.

\section{Conclusion}

We propose \textbf{TIDE-Flow}, a spike-based optical flow estimation framework that addresses the loss of fine-grained firing-time information and ambiguity in displacement matching. TIE preserves raw spike patterns, multi-scale temporal statistics, and boundary-aware interval cues before spatial compression, enabling more informative motion representations. DEE combines structured candidate-grid features with explicit displacement statistics to characterize competing correspondences, providing complementary geometric evidence for recurrent flow refinement. The PHM ablations quantify the contributions of both representations, and the full model achieves the lowest outlier rate in the main comparison at both temporal intervals.

\section*{LLM Usage Statement}

We used ChatGPT (OpenAI) for language editing of the
manuscript and supplementary material. We also used these tools
for notation checks and LaTeX formatting.

{
    \small
    \bibliographystyle{ieeenat_fullname}
    \bibliography{TIDE-Flow}
}


\end{document}